\documentclass[letterpaper, 10 pt, conference]{ieeeconf}  
\IEEEoverridecommandlockouts                               
\usepackage{bm}

\usepackage{cancel}
\usepackage{xifthen}
\newcommand\Ccancel[2][black]{\renewcommand\CancelColor{\color{#1}}\cancel{#2}}
\newcommand{\mz}[2][]{%
\ifthenelse{\isempty{#1}{}}{}{\Ccancel[cyan]{#1}}%
{\color{cyan} #2}%
}

\newcommand{\jim}[2][]{%
\ifthenelse{\isempty{#1}{}}{}{\Ccancel[olive]{#1}}%
{\color{olive} #2}%
}

\usepackage{graphicx}                       
\usepackage{graphics}                       
\usepackage{epsfig}                         
\usepackage[tight,footnotesize]{subfigure}  
\graphicspath{./pics/RAL_fig_for_revise}
\graphicspath{./pics/RAL_new_fig}
\usepackage{amssymb,amsmath}
\usepackage{mdwmath}
\usepackage{commath}   
\usepackage{eqparbox}
\usepackage{mathtools}
\usepackage[utf8]{inputenc} 
\usepackage[english]{babel}
\usepackage{amsmath,amsfonts,amssymb}

\newcommand{\ie}{{\em i.e.}}

\newcommand{\eg}{{\em e.g.}}

\newcommand{\beq}{\begin{equation}}
\newcommand{\eeq}{\end{equation}}
\newcommand{\bear}{\begin{eqnarray}}
\newcommand{\bears}{\begin{eqnarray*}}
\newcommand{\eear}{\end{eqnarray}}
\newcommand{\eears}{\end{eqnarray*}}
\newcommand{\bdm}{\begin{displaymath}}
\newcommand{\edm}{\end{displaymath}}
\newcommand{\lba}{\left[\begin{array}}
\newcommand{\ear}{\end{array}\right]}

\usepackage{stfloats}                       
\usepackage{url} 
\usepackage{hyperref}
\usepackage{cite}                           
\usepackage[T1]{fontenc} 
\usepackage{tabularx}
\usepackage{multirow}
\usepackage[table]{xcolor}
\usepackage{longtable}
\usepackage{booktabs} 			
\usepackage{xcolor,colortbl}    

\usepackage{multicol}
\usepackage{graphicx}
\usepackage{placeins}

\title{\LARGE \bf Invariant Transform Experience Replay: \\
Data Augmentation for Deep Reinforcement Learning}
\author{
Yijiong Lin$^\dagger$, 
Jiancong Huang$^\dagger$, 
Matthieu Zimmer$^\ddagger$, 
Yisheng Guan$^\dagger$,
Juan Rojas$^\mathsection$,
Paul Weng$^{\ddagger}$ \\
\thanks{$^\dagger$ Guangdong University of Technology, $^\ddagger$ Shanghai Jiao Tong University, $^\mathsection$ Chinese University of Hong Kong, Corresponding author: \url{paul.weng@sjtu.edu.cn}}
}
\begin{document}
\maketitle
\thispagestyle{empty}
\pagestyle{empty}
\begin{abstract}
Deep Reinforcement Learning (RL) is a promising approach for adaptive robot control, but its current application to robotics is currently hindered by high sample requirements.
To alleviate this issue, we propose to exploit the symmetries present in robotic tasks.
Intuitively, symmetries from observed trajectories define transformations that leave the space of feasible RL trajectories invariant and can be used to generate new feasible trajectories, which could be used for training.
Based on this data augmentation idea, we formulate a general framework, called Invariant Transform Experience Replay that we present with two techniques:
(i) Kaleidoscope Experience Replay exploits reflectional symmetries and (ii) Goal-augmented Experience Replay which takes advantage of lax goal definitions.
In the Fetch tasks from OpenAI Gym, our experimental results show significant increases in learning rates and success rates. Particularly, we attain a 13, 3, and 5 times speedup in the pushing, sliding, and pick-and-place tasks respectively in the multi-goal setting. Performance gains are also observed in similar tasks with obstacles and we successfully deployed a trained policy on a real Baxter robot. Our work demonstrates that invariant transformations
on RL trajectories are a promising methodology to speed up learning in deep RL. Code, video, and supplementary materials are available at \cite{ITER_supplement}.
\end{abstract}
\section{INTRODUCTION}\label{sec:Intro}
Deep Reinforcement Learning (RL) has demonstrated great promise in recent years \cite{mnih2015human, alphago}.
However, despite being shown to be a viable approach in robotics \cite{pmlr-v87-kalashnikov18a, openai2018learning}, deep RL still suffers from significant low sample efficiency in practice---an acute issue in robot learning.

A natural approach to deal with this issue is to better exploit the actual samples generated during learning.
Indeed, this is one of the motivations behind experience replay (ER) \cite{lin1992self} and  hindsight experience replay (HER) \cite{Andrychowicz2017HindsightReplay}.
In ER, interactions with the environment are stored in a replay buffer and can then be reused multiple times for training.
HER extends this idea to multi-task settings in order to take advantage of failed trajectories (\ie, sequences of interactions between the robot and its environment).
Its basic principle is to construct successful trajectories from failed ones by changing the unachieved (original) goals to artificial goals achieved by the failed sequences.

In this work, we advance in this direction by using symmetries to generate novel feasible artificial trajectories for training from observed ones.
We use \textit{symmetry} in its mathematical sense.
In our context, it is any transformation that leaves the space of feasible trajectories invariant.
If many such transformations are used, one observed trajectory, which is costly to collect, can cheaply produce many artificial samples for training, leading to much more efficient algorithms in terms of true samples, which is an important factor in robotics.

\begin{figure}[t]
  \centering
     \includegraphics[width=0.3\linewidth]{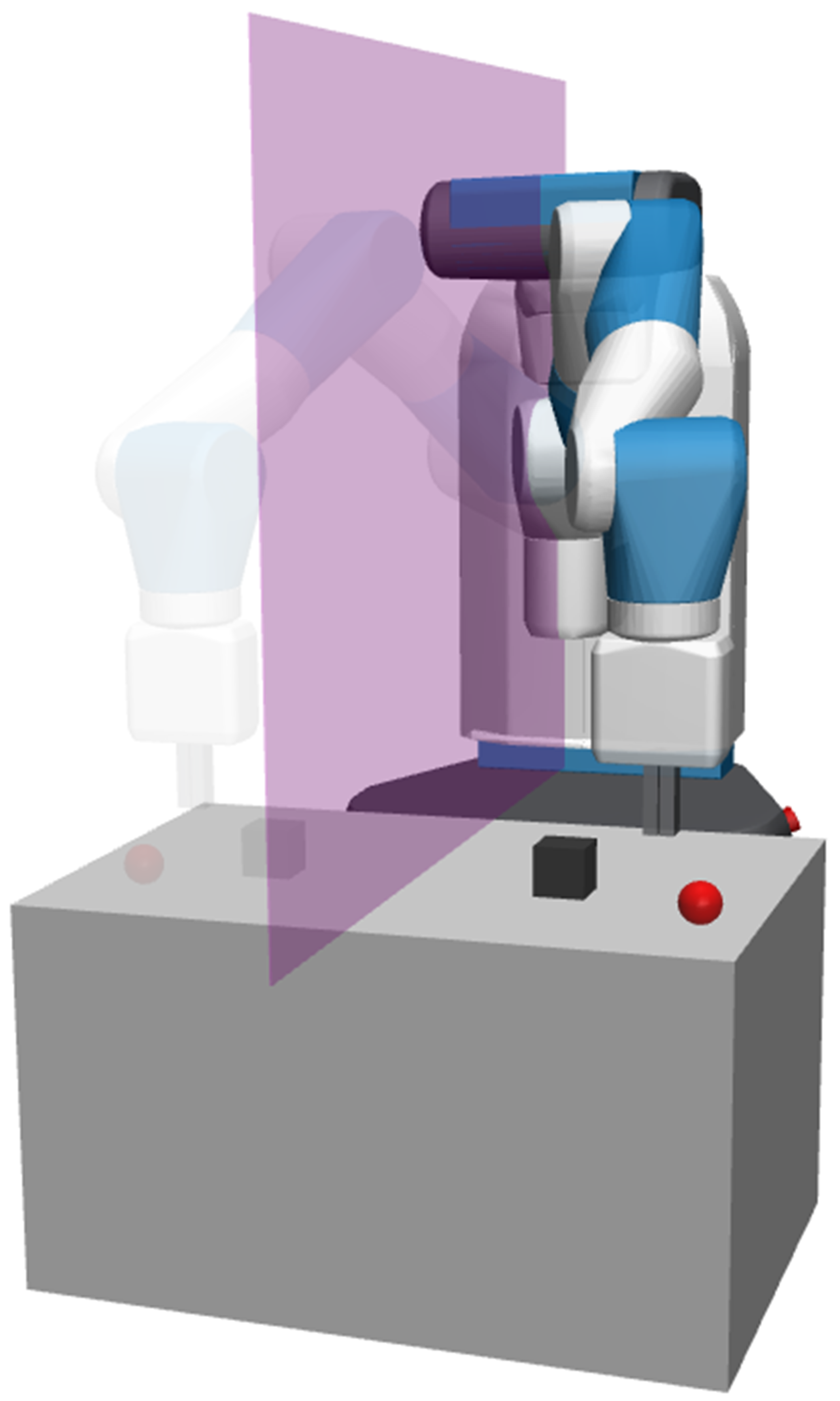}
    \includegraphics[width=.5\linewidth]{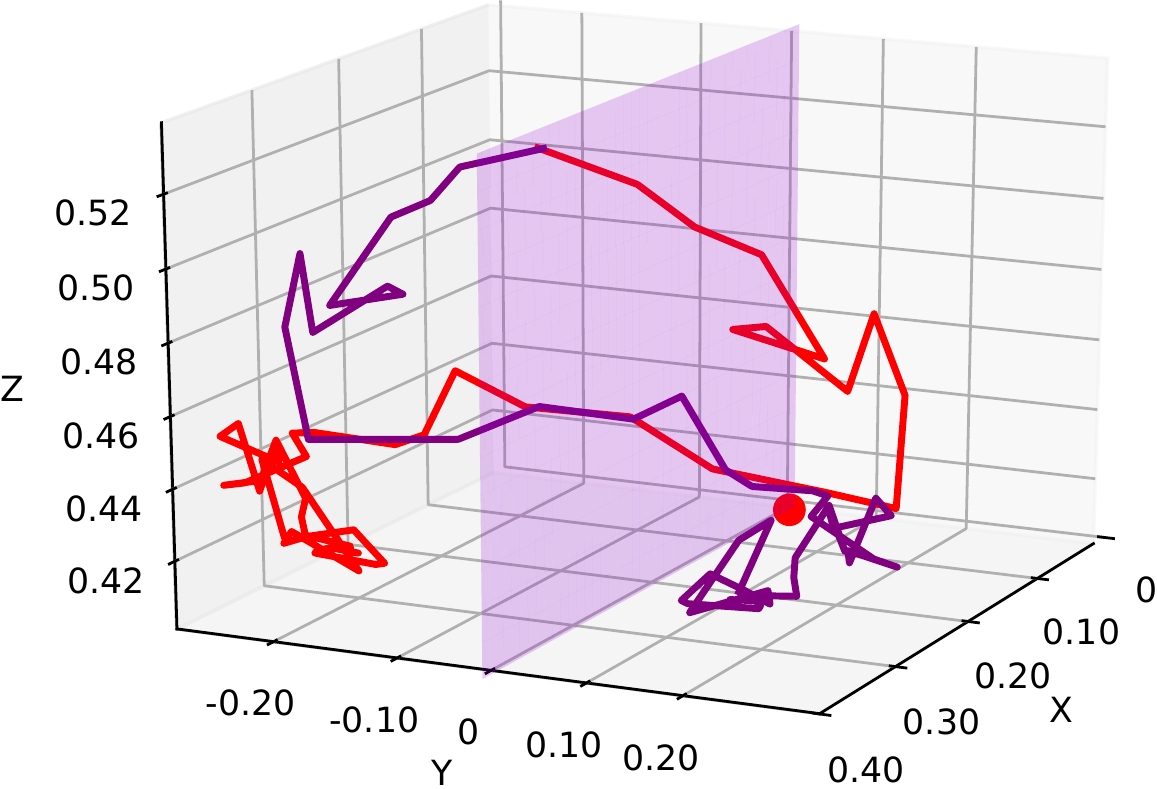}
    \caption{Left: Kaleidoscope Experience Replay leverages natural symmetry. Feasible trajectories are reflected on the plane $xoz$. The latter can itself be rotated by some $\theta_z$ along axis $\vec{z}$. 
    Right: A symmetrical trajectory (purple) is reflected from the observed trajectory (red) via the purple plane $xoz$. The red point denotes the robot base in the right plot. }
  \label{fig:ker}
\end{figure}

As a basic illustration of such transformation consider a robotic manipulation task (see the left of Fig.~\ref{fig:ker} for illustration) where the robot may interact with some objects.
The reflection with respect to the purple plane induces a transformation that maps any sequence of interactions recorded during learning to a new feasible trajectory, which can then also be used for training (see the right of Fig.~\ref{fig:ker} depicting the reflection applied to the path of the gripper).
This transformation is naturally also applied to the goal achieved by the considered trajectory (and any other state relevant information, \eg, object positions).
The intuition is that if that trajectory has achieved a goal $g$, then its transformed trajectory defines a feasible sequence of controls that achieves the symmetrical reflection of $g$.
Interestingly, such a transformation preserves any contact the robot may have with its environment if the transformation is applied to all the objects and obstacles in the robot's workspace.


The idea of using reflections and more general symmetries (see Related Work in Sec. \ref{sec:related}) to expand the original training data is the basis of many data augmentation techniques in deep learning, but has been scarcely investigated in deep RL to the best of our knowledge.

In this paper, we propose a general framework for data augmentation in deep RL, which extends ER and HER, called Invariant Transform Experience Replay (ITER) where a transformation can be applied either on trajectories entering the replay buffer or on those sampled from it.
To make ITER concrete, we present it with two different such transformations.
Each of them could potentially be used separately, leading to two independent data augmentation techniques.

The first technique, Kaleidoscope Experience Replay (KER), is based on reflectional symmetry.
It generalizes our previous example (Fig.~\ref{fig:ker}) by using multiple different reflective hyperplanes. 

The second technique, Goal-augmented Experience Replay (GER), is a direct generalization of HER:
any hindsight goal $g$ generated by HER can be instead replaced by a random goal sampled from within a small ball centered around $g$ to obtain another successful goal.
This idea takes advantage of tasks where success is defined as reaching a final pose within a distance of the goal set by a threshold (such tasks are common in robotics). 

The paper is organized as follows: Sec. \ref{sec:related} introduces related work on increasing data efficiency.
Sec. \ref{sec:back} presents the background (deep RL and HER) for our work. 
Sec. \ref{sec:aer} details our general framework with two invariant transform data augmentation techniques.
Sec. \ref{sec:expe} describes experimental results on OpenAI Gym Fetch tasks \cite{brockman2016openai}, which demonstrate the effectiveness of our propositions and show significant improvements in learning speed and success rates; particularly for robotic manipulation tasks with and without obstacles (see Fig. \ref{fig:exp_best_multigoal} and Fig. \ref{fig:exp_obstacle}).
Sec. \ref{sec:discussion} discusses concerns of interest, and Sec. \ref{sec:conclusion} concludes.
\section{RELATED WORK} \label{sec:related}

HER \cite{Andrychowicz2017HindsightReplay,Plappert2018Multi-GoalResearch} has been extended in various ways.
Prioritized replay was incorporated in HER to learn from more valuable episodes with higher priority \cite{Zhao2018Energy-BasedPrioritization}.
In \cite{Fang2019DHER:Replay}, HER was generalized to deal with dynamic goals. In \cite{Gerken2019ContinuousControllers}, a variant of HER was also investigated where completely random goals replace achieved goals and in \cite{Rauber2019HindsightGradients}, it was adapted to work with on-policy RL algorithms. All these extensions are orthogonal to our work and could easily be combined with ITER. We leave these for future work.

Symmetry has been considered in MDPs \cite{Zinkevich2001SymmetryLearning} and RL \cite{Kamal2008ReinforcementStates,Agostini2009ExploitingSpacesE,Mahajan2017SymmetryLearning,kidzinski2018learning,Amadio2019ExploitingTasks}. It can be known a priori or learned \cite{Mahajan2017SymmetryLearning}. In this work, we assume the former, which is reasonable in many robotic tasks. 
A natural approach to exploit symmetry in sequential decision-making is by aggregating states that satisfy an equivalence relation induced by some symmetry \cite{Zinkevich2001SymmetryLearning,Kamal2008ReinforcementStates}.
Another related approach takes into account symmetry in the policy representation \cite{Amadio2019ExploitingTasks}. Doing so reduces representation size and generally leads to faster solution times. However, the state-aggregated representation may be difficult to recover, especially if many symmetries are considered simultaneously.
Still another approach is to use symmetry during training instead.
One simple idea is to learn the Q-function by performing an additional symmetrical update \cite{Agostini2009ExploitingSpacesE}.
Another method is to augment the training data with their reflections \cite{kidzinski2018learning}. A dihedral group with finite invariant elements has been leveraged to implement symmetry on the state representation of board position in Go \cite{alphago}.
In this paper, we generalize further this idea and extend it to propose a general and theoretically-founded framework for data augmentation where different kinds of symmetry (not only reflections) can be considered.  

To the best of our knowledge, data augmentation has not been considered much to accelerate learning in RL. It has, however, been used extensively and with great success in machine learning \cite{Baird1992DocumentModels} and in deep learning \cite{KrizhevskyImagenetNetworks}.
Interestingly, symmetries can also be exploited in neural network architecture design \cite{Gens2014DeepNetworks}. 
However, in our case, the integration of symmetry in deep networks will be left as future work.




\section{BACKGROUND} \label{sec:back}
In this work, we consider robotic tasks that are modeled as multi-goal Markov decision processes \cite{schaul2015universal} with continuous state and action spaces: $\langle \mathcal S, \mathcal A, \mathcal G, T, R, p, \gamma \rangle$ where $\mathcal S$ is a continuous state space, $\mathcal A$ is a continuous action space, $\mathcal G$ is a set of goals, $T: \mathcal S \times \mathcal A \times \mathcal S \to [0, 1]$ is the unknown transition function that describes the environmental dynamics, $R(s, a, s', g)$ is the immediate reward when an agent reaches state $s' \in \mathcal S$ after performing action $a \in \mathcal A$ in state $s \in \mathcal S$ if the goal were $g \in \mathcal G$.
Finally, $p(s_0,g)$ is a joint probability distribution over initial states and original goals, and $\gamma \in [0, 1]$ is a discount factor.
In this framework, the robot learning problem corresponds to an RL problem that aims at obtaining a policy $\pi : \mathcal S \times \mathcal G \to \mathcal A$ such that the expected discounted sum of rewards is maximized for any given goal.

Due to the continuity of the state-action spaces, this optimization problem is usually restricted to a class of parameterized policies. 
In deep RL, the parameterization is defined by the neural network architecture.
To learn such continuous policies, actor-critic algorithms \cite{Konda1999} are efficient iterative methods since they can reduce the variance of the estimated gradient using simultaneously learned value functions.
DDPG (Deep Deterministic Policy Gradient) \cite{Lillicrap2015} is a model-free off-policy deep RL algorithm that learns a deterministic policy, which is desirable in robotic tasks. 
In DDPG, the transitions are collected into a replay buffer to later update the action-value function in a semi-gradient way and the policy with the deterministic policy gradient \cite{Silver2014}.
Because the policy has to adapt to multiple goals, as in HER, we rely on universal value functions \cite{schaul2015universal}: the classic inputs of the value function and the policy of DDPG are augmented with the desired goal.

When the reward function is sparse, as assumed here, the RL problem is particularly hard to solve.
In particular, we consider here reward functions that are described as follows:

\begin{align}\label{eq:sparserewards}
R(s, a, s', g) = \bm 1[ d(s', g) \le \epsilon_R] -1
\end{align}
where $\bm 1$ is the indicator function, $d$ is a distance (\eg, between object position in $s'$ and goal $g$), and $\epsilon_R>0$ is a fixed threshold.

To tackle this issue, HER is based on the following principle: any trajectory that failed to reach its goal still carries useful information; it has at least reached the states of its trajectory path. Using this natural and powerful idea, memory replay can be augmented with the failed trajectories by changing their goals in \textit{hindsight} and computing the new associated rewards.

In the robotic tasks solved by HER, the states are generally defined as $s = (s_{gri}, s_{obj}, s_{rel})$ with its components defined as follows: $s_{gri}$ is a 8-dimensional vector containing the absolute position of the gripper $(x_{gri},y_{gri},z_{gri})$, its linear velocity $(x'_{gri},y'_{gri},z'_{gri})$, the distance and relative velocity between the gripper's fingers $d_{fin},d'_{fin}$ respectively. Then, $s_{obj}$ is a 12-dimensional vector that consists of the pose of the object $(x_{obj}, y_{obj}, z_{obj}, \alpha_{obj}, \beta_{obj}, \gamma_{obj})$ and its twist $(x'_{obj}, y'_{obj}, z'_{obj}, \alpha'_{obj}, \beta'_{obj}, \gamma'_{obj})$. $s_{rel}$ is a 3-dimensional vector representing the position of object with respect to the target position $(x_{rel},y_{rel},z_{rel})$. 
Actions are defined as $a = (x_a, y_a, z_a, d_{gri})$ where $(x_a, y_a, z_a)$ represent the new position that the gripper should reach at the next time step and $d_{grip}$ is the desired distance between the two fingers of the gripper. Finally, goals are defined as $g = (x_g, y_g, z_g)$ specifying the target positions of objects.

\section{INVARIANT TRANSFORMATIONS FOR RL}\label{sec:aer}
To reduce the number of interactions with the real environment, we propose to generate artificial training data from observed trajectories collected during the robot's learning.
Some care is needed to choose a transformation to be applied on actual data to generate artificial ones, otherwise the training would be too biased.
To that regard, we consider symmetries (\ie, any invariant transformations) in the space of feasible trajectories.

Consider a trajectory $\tau$ of length $h$ with goal $g \in \mathcal G$ as $\langle g$, $(s_0$, $a_1$, $r_1$, $s_1$, $a_2$, $r_2$, $s_2, \ldots, s_h)\rangle$ where $s_0 \in \mathcal S$, $\forall i=1, \ldots, h$, $a_i \in \mathcal A$, $s_i \in \mathcal S$, and $r_i = R(s_{i-1}, a_i, s_i, g)$.
We assume that all trajectories have a length not larger than $H \in \mathbb N$, which is true in robotics (\ie, the length of each manipulation task is not infinite).
The set of all trajectories is denoted $\overline \Gamma = \cup_{h=1}^H \mathcal G \times \mathcal S \times (\mathcal A \times \mathbb R \times \mathcal S)^h$.

A trajectory $\tau$ is said to be \emph{feasible} if for $i=1, \ldots, h$, $T(s_{i-1}, a_i, s_i)>0$.
The set of feasible trajectories is denoted $\Gamma \subseteq \overline\Gamma$.
A trajectory $\tau$ of length $h$ with goal $g$ is said to be \emph{successful} if $R(\tau)>R_{\min}$ where $R(\tau) = \sum_{i=1}^h \gamma^{i-1} R(s_{i-1}, a_i, s_i, g)$ and $R_{\min}$ is a fixed problem-dependent threshold.
In the context of sparse rewards with $R_{\min}=0$, a successful trajectory is one that reached the goal. 
The set of successful trajectories is denoted $\Gamma^+ \subseteq \Gamma$.

We can now define the different notions of symmetries that we use in this paper.
A \textit{symmetry} of $\Gamma$ is a one-to-one mapping $\sigma : \overline\Gamma \to \overline\Gamma$ such that $\sigma(\Gamma) = \Gamma$ where $\sigma(\Gamma) = \{ \tau \in \Gamma \mid \exists \tau' \in \Gamma, \sigma(\tau') = \tau \}$.
In words, a symmetry of $\Gamma$ leaves the space invariant, \ie, it maps feasible trajectories to feasible ones.
As we only apply symmetries to feasible trajectories, we directly consider their restrictions to $\Gamma$ and keep the same notation, \ie, $\sigma : \Gamma \to \Gamma$.

A \emph{decomposable} symmetry is a symmetry $\sigma$ such that there exist one-to-one mappings $\sigma_{\mathcal G} : \mathcal G \to \mathcal G$, $\sigma_{\mathcal S} : \mathcal S \to \mathcal S$, and $\sigma_{\mathcal A} : \mathcal A \to \mathcal A$ that satisfy for any $\tau \in \Gamma$:
\begin{align}\label{eq:decomposable}
    \sigma(\tau) = \langle g', (s_0', a_1', r_1', s_1', a_2', r_2', s_2', \ldots, s_h')\rangle
\end{align}
where 
$\tau = \langle g$, $(s_0$, $a_1$, $r_1$, $s_1$, $\ldots, s_h)\rangle$,
$g' = \sigma_{\mathcal G}(g)$, 
$s'_0 = \sigma_{\mathcal S}(s_0)$,
$\forall i=1, \ldots, h$, 
$a'_i = \sigma_{\mathcal A}(a_i)$, 
$s'_i = \sigma_{\mathcal S}(s_i)$, and
$r'_i = R(\sigma_{\mathcal S}(s_{i-1}), \sigma_{\mathcal A}(a_i), \sigma_{\mathcal S}(s_i), \sigma_{\mathcal G}(g))$.
In words, a decomposable symmetry is a simple mapping that applies transformations separately on states, actions, and rewards.

A \emph{reward-preserving} symmetry $\sigma: \Gamma \to \Gamma$ is a symmetry such that for any trajectory $\tau \in \Gamma$, the rewards appearing in $\tau$ are exactly the same as those in $\sigma(\tau)$ in the same order.
In words, the value of rewards in a trajectory is not changed by a reward-preserving symmetry.
The previous definitions of symmetries can naturally be applied to the set of successful trajectories $\Gamma^+$ as well.

Besides, note that any number of symmetries induces a group structure (\ie, they can be composed).
Given $n$ symmetries and a trajectory, one could possibly generate up to $2^n-1$ new trajectories\footnote{Though in our implementation, we use $2n-1$ to avoid computational costs with an exponential increase.} (see Sec. \ref{subsec:ker} for specifics).
This property is useful if one only knows a fixed number of symmetries for a given problem, because a recursive application of those symmetries could lead to an exponential increase of trajectories that could be used for training.

\begin{figure}[t]
  \centering
    \includegraphics[width=1\linewidth]{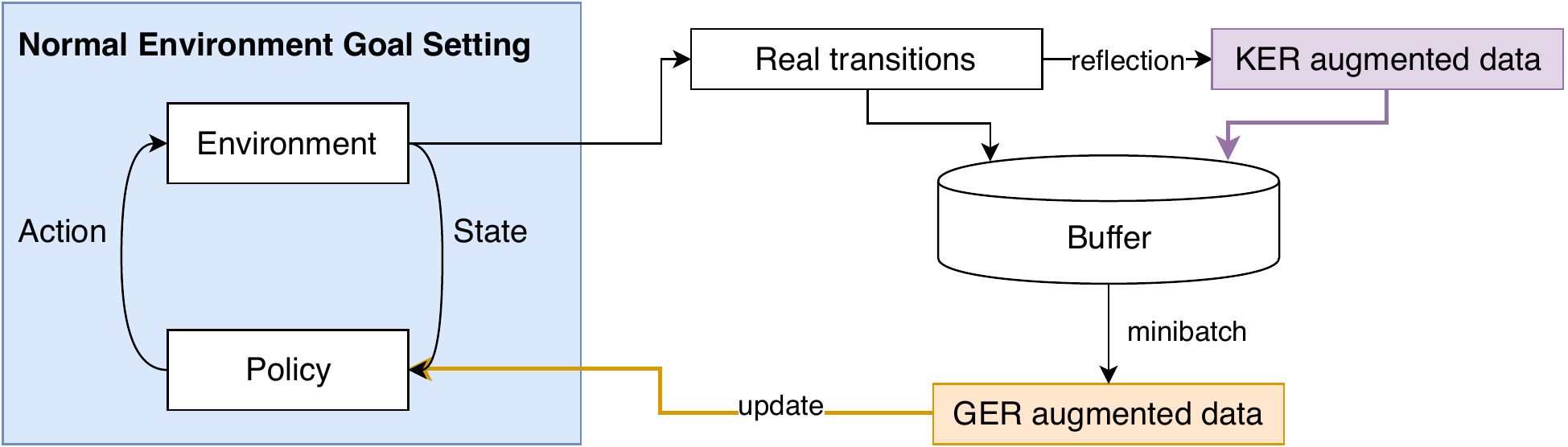}
    \caption{ITER framework overview: observed and symmetrically transformed transitions are stored in the replay buffer. 
    Sampled minibatches are then augmented with GER before updating the policy.}
  \label{fig:architecture}
\end{figure}

As a general approach to increase data efficiency in deep RL, one can leverage the symmetries of $\Gamma$ or $\Gamma^+$ for data augmentation.
As an illustration, we propose ITER (Invariant Transform Experience Replay), a general architecture for data augmentation in deep RL, which we instantiate with two techniques for concreteness:
\begin{itemize}
    \item Kaleidoscope experience replay (KER, Sec. \ref{subsec:ker}) is based on reward-preserving decomposable symmetries of $\Gamma$ and is applied to observed trajectories before they are stored in the replay buffer.
    \item Goal-Augmented Experience Replay (GER, Sec. \ref{subsec:ger}) is based on reward-preserving decomposable symmetries of $\Gamma^+$, but can be applied to all feasible trajectories in the same fashion as HER.
    It is applied to trajectories sampled from the replay buffer.
\end{itemize}
In this general framework, other symmetries (\eg, rotation, translation) could be used instead or in conjunction of KER or GER.
Besides, these two approaches are orthogonal to each other, and could be used separately. 
An overview of our architecture is illustrated in Fig. \ref{fig:architecture}.
Furthermore, note that our two methods preserve any contact that may occur between the robot and any object it may encounter (table included) as long as a symmetry is applied to all the objects and obstacles in the robot's workspace.
Therefore, our approach also works in any contact-rich robotic task, including 
problems where some obstacles may limit the movements of objects or the robot. When the poses of obstacles are given in each state but not fixed across episodes, the agent can learn the effects of contact.
For example, the agent can avoid obstacles or leverage contact to reach a goal (\eg  in the pushing task it may learn to push an object and let the obstacle stop the moving object).
\subsection{Kaleidoscope Experience Replay (KER)}\label{subsec:ker}
KER uses reflectional symmetry. 
Consider a 3D workspace with a bisecting plane $xoz$ as shown in Fig. \ref{fig:ker}. If a trajectory is observed in the workspace (red in Fig. \ref{fig:ker}), the symmetry  associated to $xoz$ would then yield a new feasible trajectory reflected on this plane. More generally, the $xoz$ plane may be rotated by some angle $\theta_z$ along axis $\vec{z}$ and still define an invariant symmetry for the robotic task.

\begin{figure}[t]
  \centering
    \includegraphics[width=1\linewidth]{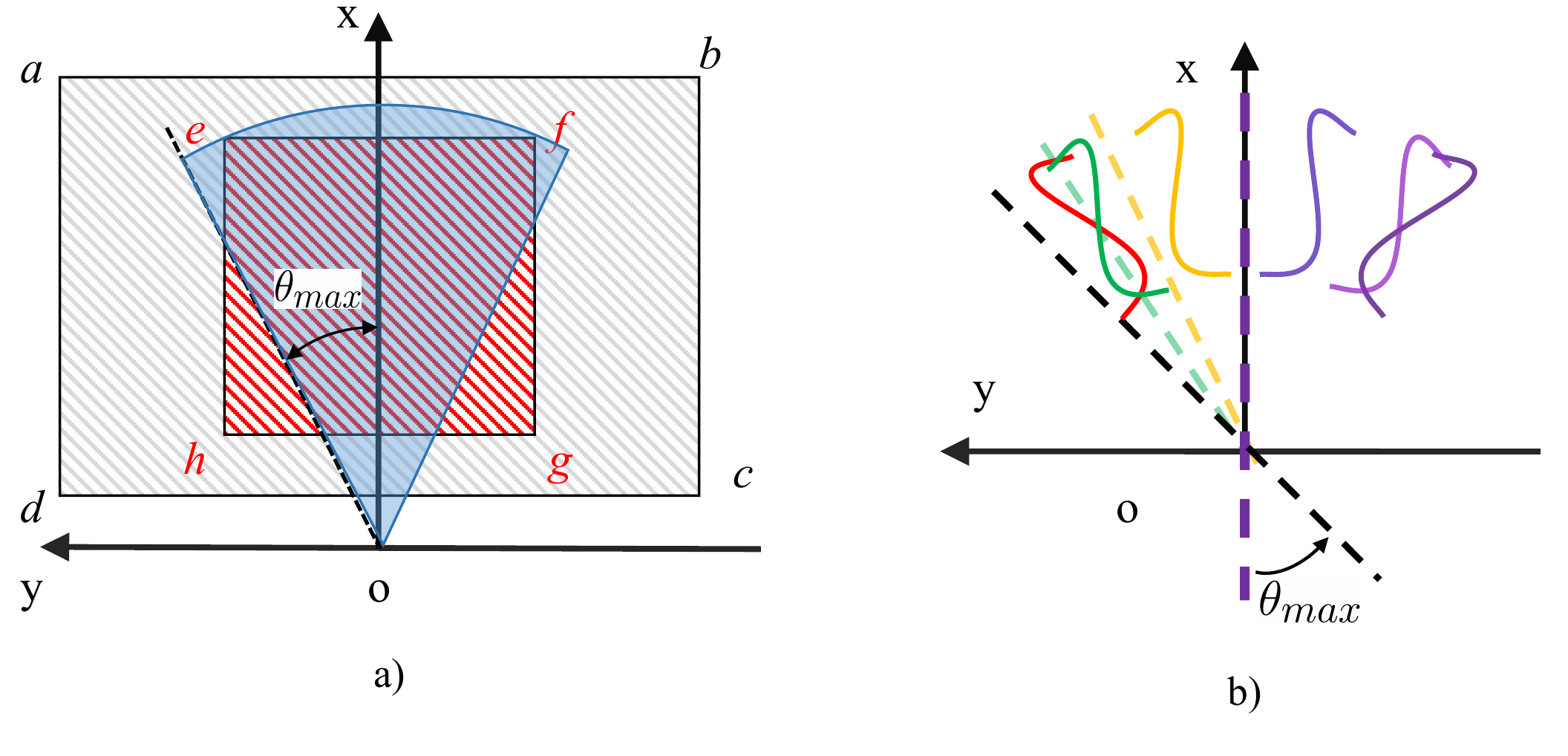}
    \caption{a) Grey represents the valid workspace (the table surface). Red represents the object and goal's possible initial positions. Blue represents valid areas for symmetry hyperplanes (lines in this 2-dimensional visualization) that KER applies to reflect any observed trajectory. b) 2D illustration of how KER reflects an observed trajectory when $n_\text{KER}=3$. 
    }
  \label{fig:theta_max}
\end{figure}

We can now precisely define KER, which amounts to augmenting any observed trajectory with a certain number of random decomposable symmetries.
To generate feasible symmetrical trajectories, one must choose a maximum valid angle $\theta_\text{max}$ for generating symmetries in any specific robotic manipulation task as shown in Fig. \ref{fig:theta_max} a). 
Value $\theta_\text{max}$ is a hyperparameter that one can enlarge to expand the number of symmetrical trajectories (leading to more general policy training). Note, however, that it is also possible that a limited number of reflections lead to trajectories that consists of sections where the robot manipulator is outside the workspace. In these cases, such trajectories are not included in the replay buffer to ensure that the reflection is invariant.

State, action, and goal vectors consist of position, orientation, linear velocity, and angular velocity elements that can be reflected through symmetry. Position and linear velocity are represented through variables $(x,y,z)$, whilst variables $(\alpha, \beta, \gamma)$ are used to represent the orientation and angular velocity elements (as recalled in Sec. \ref{sec:back}). The center of the robot's base coincides with the origin. Whenever, a robot's shoulder is offset from the origin and we need to consider the parallel sagittal plane, then we translate all coordinates  to this new plane. Since the distance between fingers and their corresponding relative velocity are scalar, they remain unchanged after symmetry. Thus, we only consider reflecting $(x,y,z)$ and $(\alpha, \beta, \gamma)$ to KER-augmented elements $(x_{sym},y_{sym},z_{sym})$ and $(\alpha_{sym}, \beta_{sym}, \gamma_{sym})$. 
Note that each plane in the 3-dimensional Cartesian space can be leveraged to yield one symmetry. Formally, each plane $\psi$ is only associated with one symmetry $\sigma^\psi$.
The number of reflections used in KER is controlled by hyperparameter $n_\text{KER}$. 
If $n_\text{KER} =1$, KER directly applies $\sigma^{xoz}$ to reflect the new trajectory in relation to states, actions, and goals in terms of $(x,y,z)$ and $(\alpha, \beta, \gamma)$ elements as shown below:
\begin{equation}\label{eq:ker_1_pos}
	\begin{split}
    \sigma^{xoz}((x,y,z))&= (x,-y,z) \\
     &=(x_{sym},y_{sym},z_{sym})\\
	\end{split}
\end{equation}
\begin{equation}\label{eq:ker_1_ori}
	\begin{split}
	\sigma^{xoz}((\alpha,\beta,\gamma))\\
     &=(-\alpha,\beta,-\gamma)\\
     &=(\alpha_{sym}, \beta_{sym}, \gamma_{sym})
	\end{split}
\end{equation}
If $n_{\text{KER}} >1$, there are two stages. In the first stage, 
KER generates a set of rotated symmetric planes $\Psi =\{ \psi_{\theta_j}^{z} \mid \theta_j\in \Theta  \}$ which are rotated along the $\vec{z}$-axis by a set of uniformly sampled angles $\Theta =  \{ \theta_j \mid \theta_j \sim (0,\theta_\text{max}], j=1,2,...,n_{\text{KER}}-1  \}$.
The set of associated decomposable symmetries for those planes is denoted as $\Sigma^{\Psi} =\{  \sigma^{\psi_{\theta_j}} : \Gamma \to \Gamma \mid \psi \in \Psi, j = 1,2,...,n_{\text{KER}}-1 \}$ (here for conciseness we use $\psi_{\theta_j}$ to represent $\psi_{\theta_j}^{z}$).
The decomposition of $\sigma^{\psi_{\theta_j}}$ is:
\begin{equation}
	\begin{split}
    \sigma^{\psi_{\theta_j}}((x,y,z))&=  Rot_z(\theta_j) [\sigma^{xoz}(Rot^{-1}_z(\theta_j)  (x,y,z)^{T})]^{T}\\
     &=(x_{sym},y_{sym},z_{sym})
	\end{split}
\end{equation}
\begin{equation}
	\begin{split}
\sigma^{\psi_{\theta_j}}((\alpha,\beta,\gamma))&= 
Car(Rot_z(\theta_j)Eul(\dot{\alpha},\dot{\beta},\dot{\gamma})) \\
&= (\alpha_{sym},\beta_{sym},\gamma_{sym})\\
	\end{split}
\end{equation}
where:
\begin{align*}
(\dot{\alpha},\dot{\beta},\dot{\gamma}) &= \sigma^{xoz}(Car(Rot^{-1}_z(\theta_j)  Eul(\alpha,\beta,\gamma)))
\end{align*}
with $Eul:\mathbb{R}^{3}\rightarrow SO(3)$ which maps a 3D-vector $(\alpha, \beta, \gamma)$ of Euler angles to a rotation matrix $Rot \in SO(3) \subset \mathbb{R}^{3\times 3} $ (where $SO(n)$ denotes the special orthogonal group of dimension $n$) under right-handed coordinate frame, and $Car: SO(3)\rightarrow \mathbb{R}^{3}$ is the inverse mapping of $Eul$ following the  $x$-, $y$-, and $z$-axes rotation sequence (Cardano sequence)\cite{Corke2017Robotics}. $Rot_z(\theta)$ is a standard rotation matrix for rotation of $\theta$ about the $z$ axis.
At the end of the first stage, we obtain a set of trajectories $\overline{\Gamma}_{1}$ consisting of an observed trajectory and its symmetrical trajectories generated from $\Sigma^{\Psi}$. 
From this set, infeasible trajectories (\eg, out of workspace) are filtered out to define a set of feasible trajectories ${\Gamma_{1}}$.
Then KER applies $\sigma^{xoz}$ to ${\Gamma_{1}} $ according to Eqn. \ref{eq:ker_1_pos} and Eqn. \ref{eq:ker_1_ori}, and then yields another set of feasible symmetrical trajectories ${\Gamma_{1}'} $ (all these trajectories are feasible since our workspace is symmetrical with respect to $xoz$ plane). Finally, the trajectories in the set ${\Gamma_{2}} = {\Gamma_{1}'}  \cup {\Gamma_{1}} $ are stored into the replay buffer in each episode.
In general, the maximum number of new trajectories generated is $ 2 n_{\text{KER}} - 1$. Consider Fig. \ref{fig:theta_max} b), here KER first samples two symmetry hyperplanes (the green and yellow lines) rotated about the $z$-axis with origin $o$ with uniformly sampled angles with range $(0,\theta_{max}]$. Then, we reflect the observed (red) trajectory across the two planes to generate two new trajectories (yellow and green). Finally, KER reflects both the observed and reflected trajectories about the $x$-axis to generate three new purple trajectories.
 
Note that instead of storing the reflected trajectories in the replay buffer, random symmetries can also be applied to sampled minibatches from the buffer. This approach was tried previously for single-symmetry scenarios \cite{kidzinski2018learning}. However, the approach is more computationally taxing (as transitions are reflected every time they are sampled) and leads to lower performance, which is due to a lower diversity in the minibatches as discussed in Sec. \ref{sec:discussion}.
\subsection{Goal-Augmented Experience Replay (GER)}\label{subsec:ger}
\begin{figure}[t]
  \centering
      \includegraphics[width=1.\linewidth]{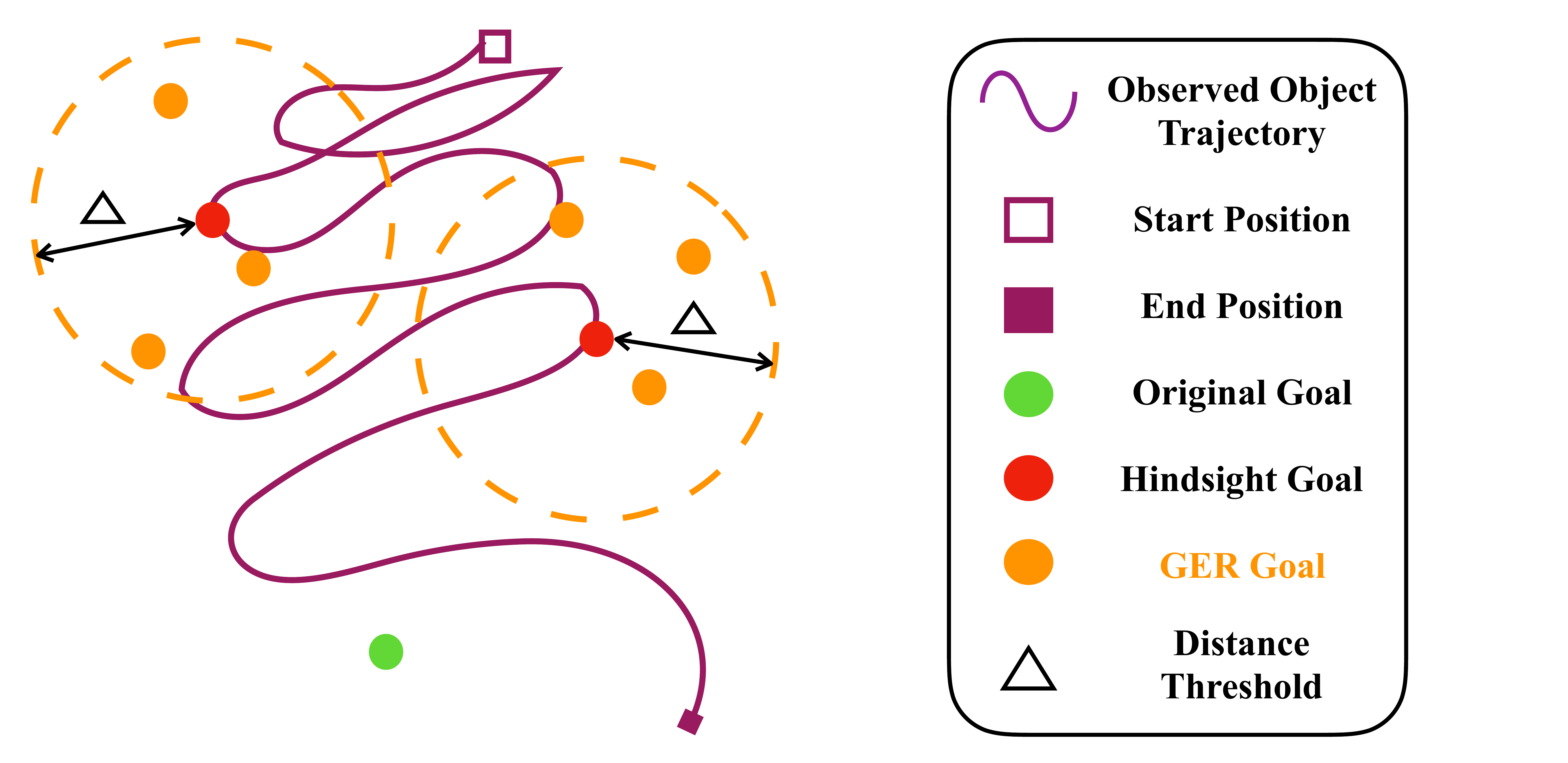}
    \caption{An illustration for GER.}
  \label{fig:ger_2d}
\end{figure}
GER exploits any reward function formulation (see Eqn.~\ref{eq:sparserewards}) that defines a successful trajectory as one whose end position is within a small radial threshold (a ball) centered around the goal. Thus, when the robot obtains a trajectory, we can consider it successful for any goal within a ball centered at each state of that trajectory.
Based on this observation, GER augments trajectories by replacing the original goal with a random goal sampled uniformly within that ball. Here is an example in pushing task. As shown in Fig. \ref{fig:ger_2d}, when HER chooses some hindsight goals (red points) to replace the original goal (green point) for experience replay, we can further sample more goals (orange points) within the circle with radius $\Delta$ that also satisfy the success condition for hindsight goal replacements, and we call those artificial goals (red and orange points) as GER-goals.  
This ball can be formally described as $B(s_h, \Delta) = \{ g \in \mathcal G \mid d(s_h, g) \le \Delta  \}$ where $s_h$ is the state reached in the observed trajectory and $\Delta \le \epsilon_R$ is a threshold.

Formally, GER is based on reward-preserving decomposable symmetries of $\Gamma^+$ where $\sigma_{\mathcal S}$ and $\sigma_{\mathcal A}$ are identity mappings and $\sigma_{G}$ is randomly chosen, conditional to a trajectory $\tau$ reaching some state $s_h$, in the following set:
$\{ \rho : \mathcal G \to \mathcal G \mid \forall g \in \mathcal G, \mbox{ }\rho(g) \in B(s_h, \Delta)\}$.
Interestingly, such symmetries, when viewed as mappings from $\Gamma$ to $\Gamma^+$ can be applied to the whole set of feasible trajectories to generate successful trajectories, which we do in our architecture.
In this sense, GER is a generalization of HER and can be implemented in the same fashion. $n_\text{GER}$ is a hyperparameter that controls the ratio between the numbers of original goals $g\in \mathcal G_{Ori}$ and GER goals $g \in \mathcal G_{GER}$ used in minibatch for policy training: $\frac{\left | \mathcal G_{GER} \right |}{\left | \mathcal G_{Ori} \right |}=n_\text{GER}$. Note that in order to take full advantage of realized goals, in our definition, when $n_\text{GER}=1$, HER is a spacial form of GER with $\Delta=0$, which means the GER-goals are all the hindsight goals chose right on the state of any observed trajectories.

\section{Experiments and Results} \label{sec:expe}
\begin{figure*}[h]
  \centering
    \includegraphics[width=1.\textwidth]{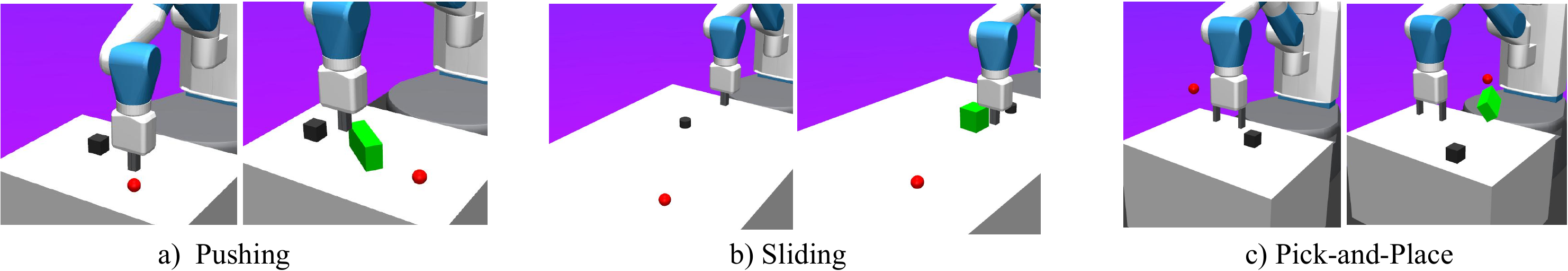}
\caption{Evaluation on robotic tasks without obstacles \cite{Andrychowicz2017HindsightReplay} (left) and with obstacles (showed as green bricks in right) . Goals are represented by red balls.}
  \label{fig:tasks}
\end{figure*}

\begin{figure}[h]
  \centering
    \includegraphics[width=.48\textwidth]{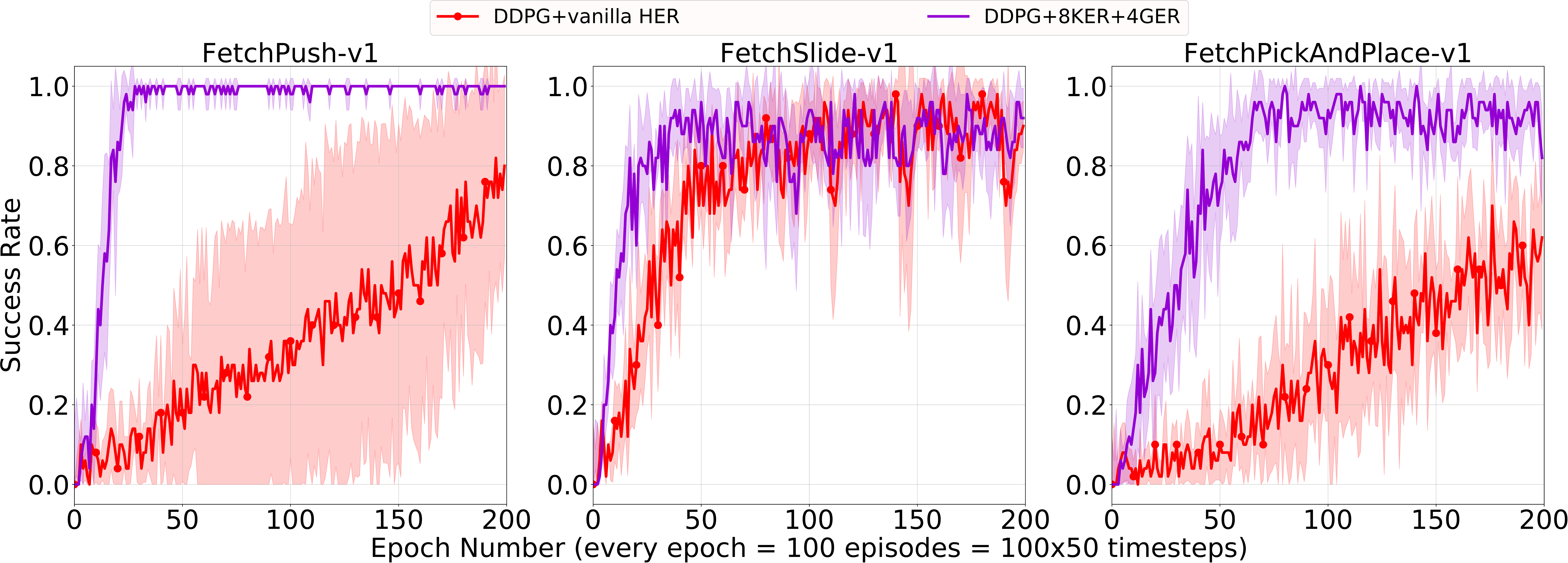}
\caption{Comparison of vanilla HER and ITER with 8 KER symmetries and 4 GER applications on obstacle-free tasks.}
  \label{fig:exp_best_multigoal}
\end{figure}
\begin{figure}[h]
  \centering
    \includegraphics[width=0.48\textwidth]{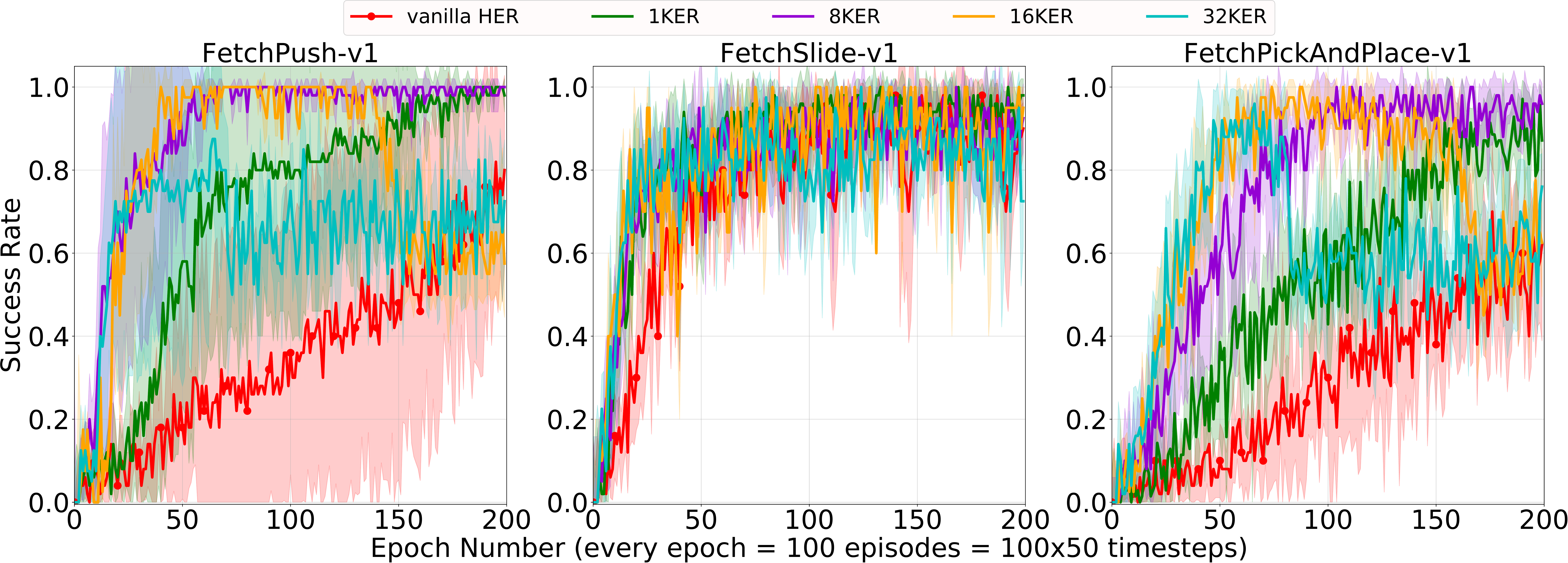}
\caption{Comparison of different $n_{\text{KER}}$ for KER with a single GER on obstacle-free tasks.}
\label{fig:exp_nker}
\end{figure}
\begin{figure}[h]
  \centering
    \includegraphics[width=0.48\textwidth]{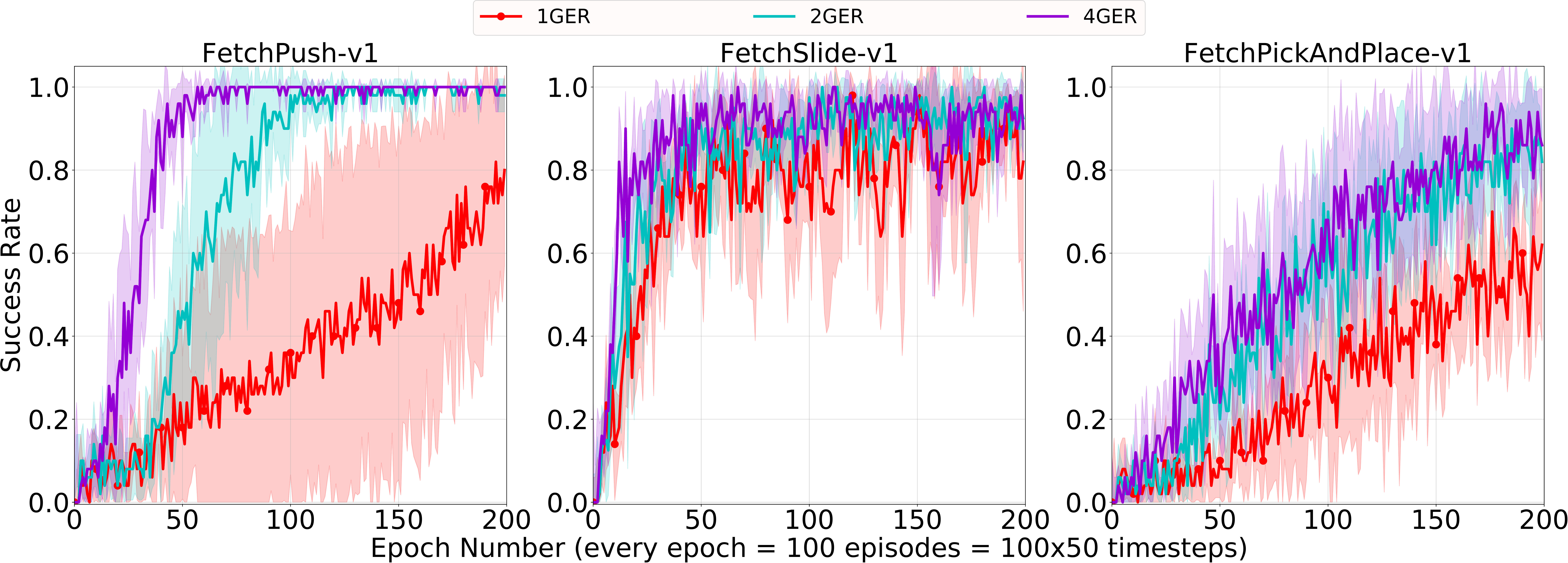}
\caption{Comparison of different $n_{\text{GER}}$ for GER without KER on obstacle-free tasks.}
\label{fig:exp_nger}
\end{figure}
\begin{figure}[h]
  \centering
    \includegraphics[width=0.48\textwidth]{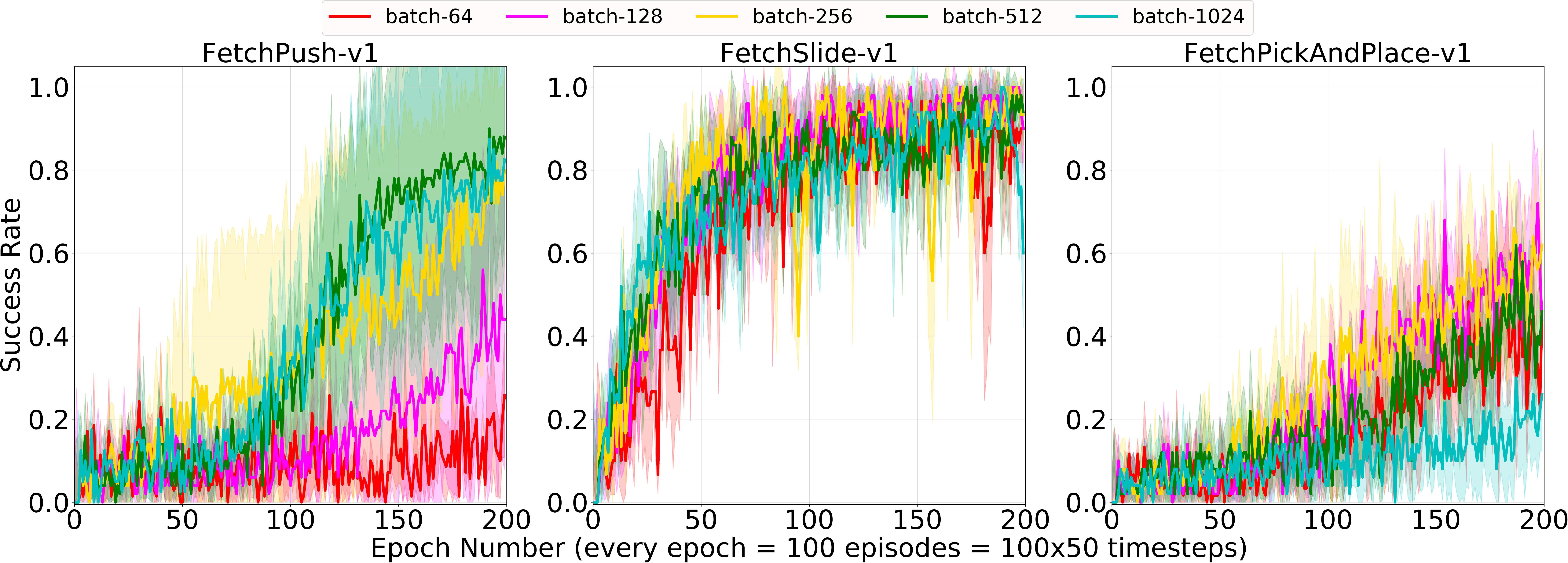}
\caption{Comparison of vanilla HER with different minibatch sizes from 64 to 1024.}
\label{fig:exp_her_batch_size}
\end{figure}
\begin{figure}[h]
    \includegraphics[width=0.48\textwidth]{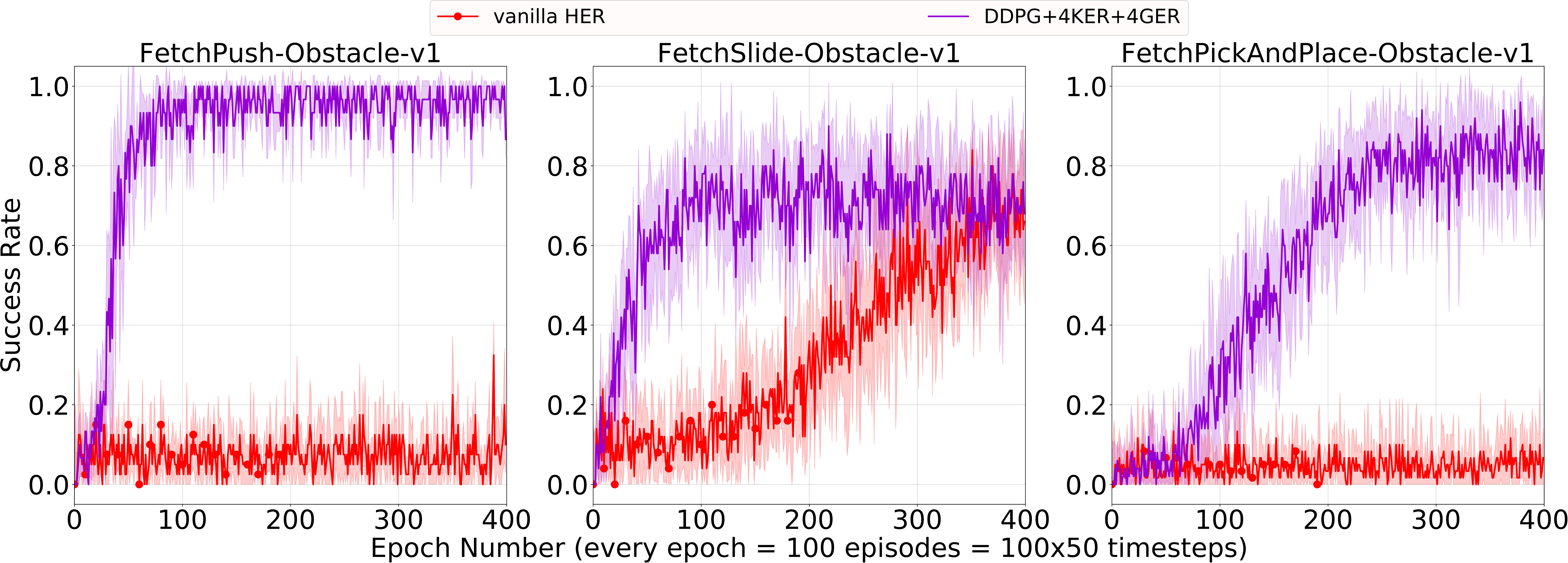}
\caption{Comparison of vanilla HER and ITER with 4 KER symmetries and 4 GER applications on tasks with obstacles.}
\label{fig:exp_obstacle}
\end{figure}

\subsection{Environment}
To evaluate our method, a simulated 7-DOF (degrees of freedom) Fetch arm with a two-fingered parallel gripper is trained with DDPG on the pushing, sliding, and pick-and-place tasks from OpenAI Gym \cite{brockman2016openai}. The state and action are defined according to Sec. \ref{sec:back}. 
The rewards are sparse and binary. If successful (the goal is achieved within an error distance) the agent gains a 0 reward, otherwise -1. With regards to goal replay, \cite{Andrychowicz2017HindsightReplay} proposed four strategies for selecting the replayed goal. Their experimental results show that the \textit{future} strategy performed best. As such, we use the same goal sampling strategy in all of our experiments. In our case, this strategy will select $k$ random states $s_h$ (that will be set at the center of the ball for sampling random goals in GER experiments as shown in Fig. \ref{fig:ger_2d}) that come from the same episode as the transition being replayed and were observed afterwards. In our experiment, we use the same $k$ as in \cite{Andrychowicz2017HindsightReplay}, which is 8. 

Our method is evaluated on three simulated manipulation 
tasks described below (and introduced in \cite{Andrychowicz2017HindsightReplay}) and shown in Fig. \ref{fig:tasks}. For all tasks, a movable object is initialized randomly on a table.  
\subsubsection{Pushing} The robot's aim is to move the object to a desired position on the table.
\subsubsection{Sliding} The robot's aim is to slide the object to a goal position on the table (the goal position is outside the robot’s workspace). The robot must learn to contact the object with enough momentum such that it reaches its goal (considering friction).
\subsubsection{Pick-and-place} The robot's aim is to move the object to a desired position in space. 

Note that for the pushing and sliding tasks, the fingers are blocked to prevent the agent from learning to grasp. All hyperparameter values are set equal to those presented in \cite{Andrychowicz2017HindsightReplay}.

\textit{Learning with Obstacles}:
Our method also succeeds in more complex environments; namely, those with obstacles. In each episode, a static brick-like obstacle is randomly placed in the robot workspace (see Appendix A \cite{ITER_supplement} for details). 
The state space dimensionality increases to 31 and additionally includes the obstacle pose.
In such scenarios, the robot must learn how to manipulate a movable object to achieve a goal by possible interactions with the obstacle---a much harder learning process.
\subsection{Training Setting}
The training setting
is conducted according to \cite{Andrychowicz2017HindsightReplay}. Hyperparameter values are unchanged unless otherwise stated. We train policies on a single machine with 1 CPU core and generate experiences by using 2 rollouts. 

An epoch is defined as a fixed-size set of successive episodes. Since a trajectory (also defined as a single episode) can be considered \textit{successful} (Sec. \ref{sec:aer}), we can compute the \textit{success rate} over an epoch by counting the number of successful episodes. 
To highlight the difference in learning rate ability between ITER and HER, we use an eighth of the number of episodes (100 episodes per epoch instead of 800) in each training epoch.

\begin{figure}[h]
  \centering
    \includegraphics[width=0.48\textwidth]{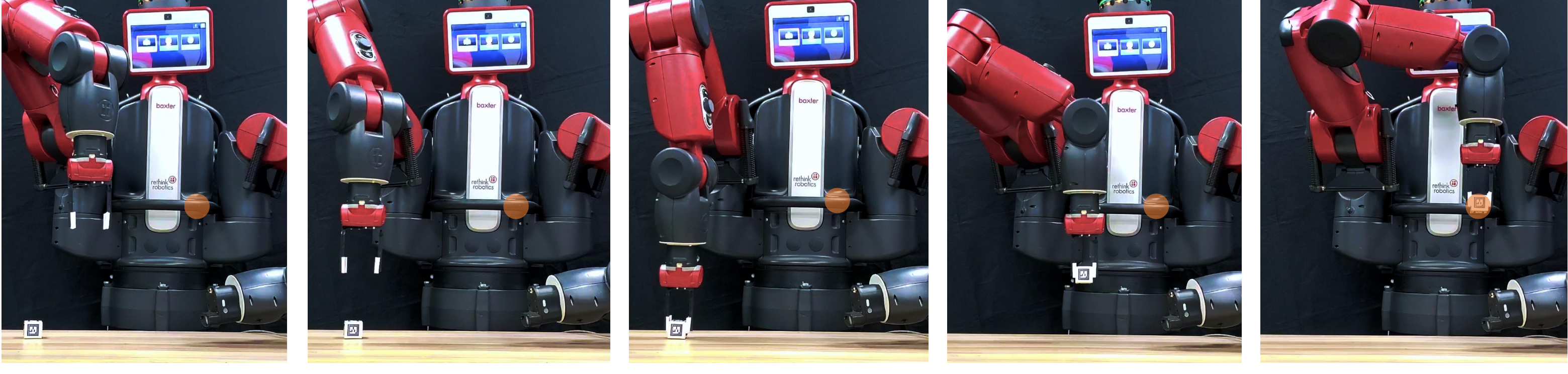}
\caption{A real Baxter robot running a pick-and-place policy trained via ITER (the goal is located at the orange ball).}
\label{fig:exp_real_baxter}
\end{figure}

Note that during learning, the discount factor, the structures of the policy network and the Q-value network (input, output, activation functions, and hidden layers), the policy's exploration strategy, the optimization method, the learning rates, the soft update ratio, and the replay buffer size are all equal to \cite{Andrychowicz2017HindsightReplay}.
Policy performance during testing is shown in Figs. \ref{fig:exp_best_multigoal}-\ref{fig:exp_obstacle}. That is, exploration is disabled, rendering the policy fully deterministic. After each learning epoch, the testing success rate is computed over 10 episodes. We display an average over 5 random seeds for each curve.

Finally, note that a successful episode is defined as having an object reach a final position within distance $\epsilon_R$ of the goal. Namely, 5cm for pushing and pick-and-place, and 20cm for sliding. 
To sample the GER goals, we used 2D-balls in the pushing and the sliding tasks, and 3D-balls in the pick-and-place task ($\Delta$ equals $\epsilon_R$).

We design experiments to demonstrate the effectiveness of our approach and to answer the following questions.
\begin{itemize}
    \item How does ITER (GER+KER) perform compared to HER on obstacle-free robotic tasks?
    \item How much does KER contribute to ITER's performance? How many $n_\text{KER}$ should be used?
    \item What is the contribution of GER to the performance of ITER? What is the impact of $n_\text{GER}$ ?
    \item Does ITER (GER+KER) improve performance even with an obstacle in the robot workspace?
    \item Could we deploy a well-trained policy learned from ITER to the real robot without any finetuning?
\end{itemize}
\subsection{Does ITER improve performance with respect to HER?}
Experimental results show that when ITER uses $n_{\text{KER}}=8$ and $n_{\text{GER}}=4$ it significantly outperforms the data efficiency of HER across tasks in obstacle-free tasks (Fig. \ref{fig:exp_best_multigoal}). In this experiment, we achieve a 13$\times$, 3$\times$, and 5$\times$ speedup over HER for pushing, sliding, and pick-and-place tasks respectively. Note that the sliding task is very challenging as it is only determined by a few contacts (generally one) between the gripper and the object. The limited number of contacts limits the performance gain of ITER over HER. 

\subsection{How many symmetries should we use in KER?}
In this experiment, only a single GER application with a zero threshold $\Delta$ is used (\ie, HER).
We observe a monotonic performance increase with respect to the number of random symmetries $n_\text{KER}$ as illustrated in Fig. \ref{fig:exp_nker}.
We also note that there are performance drops for larger $n_\text{KER}$ and we present hypotheses in Sec. \ref{sec:discussion} to explain the phenomena (see Appendix B \cite{ITER_supplement} for experiments validating these hypotheses).
\subsection{Does GER improve performance?}
In this experiment, KER is not used and we only vary the number of GER applications.
As with KER, performance improves as more GERs are applied until a ceiling is reached. Results are shown in Fig. \ref{fig:exp_nger}.

Given that GER changes the size of the minibatch, we performed a controlled experiment where we increased the size of the minibatch in vanilla HER. More specifically, the size of the minibatch is set to $256*n_\text{GER}$. Theoretically, with stochastic gradient descent, learning may speed up as the size of the minibatch increases, since the approximation of the minibatch gradient is more precise according to the Law of Large Numbers. However, we found that enlarging the minibatch size in HER did not always improve the performance in the aforementioned tasks---some times even deteriorated it (Fig. \ref{fig:exp_her_batch_size}). 
We believe that with larger minibatches the network converges to sharp minimizers leading to poorer generalization performance  \cite{keskar2016largebatch}. 
In contrast, GER does not suffer such a degradation as it augments the data by introducing some noise. GER improves the learning performance despite a larger minibatch size. By observing Figs. \ref{fig:exp_nker} and \ref{fig:exp_nger}, we can conclude that KER and GER both contribute to ITER in similar proportions.

\subsection{How does ITER perform in tasks with obstacles?}
The experimental results shown in Fig. \ref{fig:exp_obstacle} manifest that HER cannot resolve pushing and pick-and-place tasks within 400 epochs due to the more complex dynamics introduced by obstacles. In contrast, ITER yielded highly efficient learning, converging to a satisfying performance in around 80 and 230 epochs respectively. 
In the sliding task, ITER converges in around 100 epochs whilst HER converges after 400 epochs. 
These experiments prove the effectiveness of our method even in contact-rich environments (videos available in \cite{ITER_supplement}). 

\subsection{Real robot deployment}
Similarly to HER \cite{Andrychowicz2017HindsightReplay}, we show that a policy trained with ITER can be transferred to a real robot.
A Rethink Baxter dual-armed humanoid was used for evaluation. First, ITER trained policies in a MuJoCo simulation with Baxter. Then we directly applied a well-trained policy from simulations to the real Baxter without fine-tuning. Object poses were detected by Alvar markers  (see Appendix C \cite{ITER_supplement} for learning plots and other details).
The real Baxter successfully achieved pick-and-place in 46 out of 50 trials as shown in Fig. \ref{fig:exp_real_baxter} (videos available in \cite{ITER_supplement}). 

\section{DISCUSSION} \label{sec:discussion}

\subsection{Transformations to the replay buffer's input or output?}
One interesting question concerns how the new data should be used. 
We could either populate the replay buffer with the artificial transitions or apply the transformations to a minibatch sampled from the replay buffer.
If the transformations are applied after, the diversity of the minibatch could be limited because all the new artificial transitions come from the same source.
On the other hand, if the transformations are applied before, then we do not fully exploit the information contained in this transformation because we only sample observed states within the same trajectory for several times.
In our experiment, we notice that applying KER before and GER after works better in practice.
\subsection{Performance drop with KER}
In the KER experiments, we noticed an unexpected performance dropped after running around certain numbers of epochs. This drop showed up earlier as the number of symmetries $n_\text{KER}$ increased. We first thought the algorithm was overfitting the new artificial goals at the expense of the real goals. However, in an experiment not shown here (see Appendix B \cite{ITER_supplement} for details), we observed that the performance drop occurs also with HER, regardless of whether ITER is used or not. DDPG seems to suffer from this instability after seeing a certain amount of data (actual or artificial).
%
\section{CONCLUSIONS} \label{sec:conclusion}
We proposed ITER, a general framework for data augmentation in deep RL, which we instantiated with two novel techniques KER and GER in both simple and complex dynamical environments. 
KER exploited reflectional symmetry in the feasible workspace while creating invariant RL trajectories. 
GER, as an extension of HER, is specific to goal-oriented tasks where success is defined with a threshold distance and generalizes hindsight goals.
These techniques greatly accelerate learning and improved success rates as demonstrated in our experiments.
As mentioned before, ITER could be formulated with other kinds of transformations (\eg, translation, rotation) as long as they satisfy the properties (\eg, reward-preserving symmetries on feasible trajectories) we introduced.
We leave the investigation of such other symmetries in ITER for future work.
ITER's accelerated learning enabled satisfactory learning performance in only 250k timesteps (2 hours of interaction time) for pick-and-place--a 500\% improvement compared to HER's. And we presented that policies learned under ITER can also endow the real robot with the pick-and-place ability without further finetuning. ITER also resolved environments with obstacles in a highly-efficient manner, while vanilla HER fails to solve some tasks. 
\section*{ACKNOWLEDGMENT}
This work is supported by GD Dept. of Science \& Tech. [2019A050510040], by the NSF of China [61950410758, 61750110521, 61872238], and the Shanghai NSF [19ZR1426700].
\bibliographystyle{IEEEtran}
\bibliography{IEEEabrv,references_jim,manual}
\pagebreak
\appendices
\section{Experiments Setting Details}\label{appendix:exp_setuZ}
In this appendix, more details on the robot manipulation environments setup are provided. 

\begin{itemize}

\item {\bfseries Environments without Obstacle:} The robotic manipulation tasks without obstacle are set exactly the same as how \cite{Andrychowicz2017HindsightReplay} did.

\item {\bfseries Environments with Obstacle:} Based on the aforementioned tasks, a static brick-like obstacle is randomly placed in the robot workspace. Specifically, in pushing and sliding tasks, bricks are placed with uniformly random pose (denoted by $(x_{obs}, y_{obs}, z_{obs}, \alpha_{obs}, \beta_{obs}, \gamma_{obs})$) on the table in the $30\times30$cm square with the center under the initial position of the robot's end-effector, and for pick-and-place task a brick's x and y coordinates are sampled from the same square and its z coordinate is sampled uniformly from $[0,25cm]$. The brick size ($length \times width \times height$) in pushing and pick-and-place tasks is $16 \times 5 \times 8$cm while the one in sliding task is $8 \times 8 \times 8$cm. Video of training on these environments is available in \cite{ITER_supplement}. 

\end{itemize}

\section{Hypotheses and Experiments on Performance Drop}\label{appendix:drop}
To systematically analyze the performance drop met in KER ablative experiment, hypotheses listed below are proposed and relative experiments are conducted to verify. For simplicity, we conducted experiments only in the pick-and-place task, and each plot of learning performance is averaged with 2 runs in this section.

\begin{itemize}

\item {\bfseries Hypothesis 1: Overfitting towards Buffer Data}\\
In deep learning, the network tends to overfit towards the training dataset after a long-run learning, leading to poorer generalization. Does it related to the performance drop here? To answer this question, we design an experiment with a new evaluation on the robot learning performance, in which goals are sampled from buffer (Fig. \ref{fig:diagram_overfit}). If it were the case of overfitting towards buffer, there should be no drop in performance. However, the experimental results in Fig. \ref{fig:plot_overfit_pnp} show that the drops are still occurred in around same epochs (80 epochs and 150 epochs for $n_{ker}=16$ and $n_{ker}=32$ respectively) as before. Therefore, the drop is not caused by overfitting towards buffer data.
\begin{figure*}
  \centering
        \includegraphics[width=0.8\linewidth]{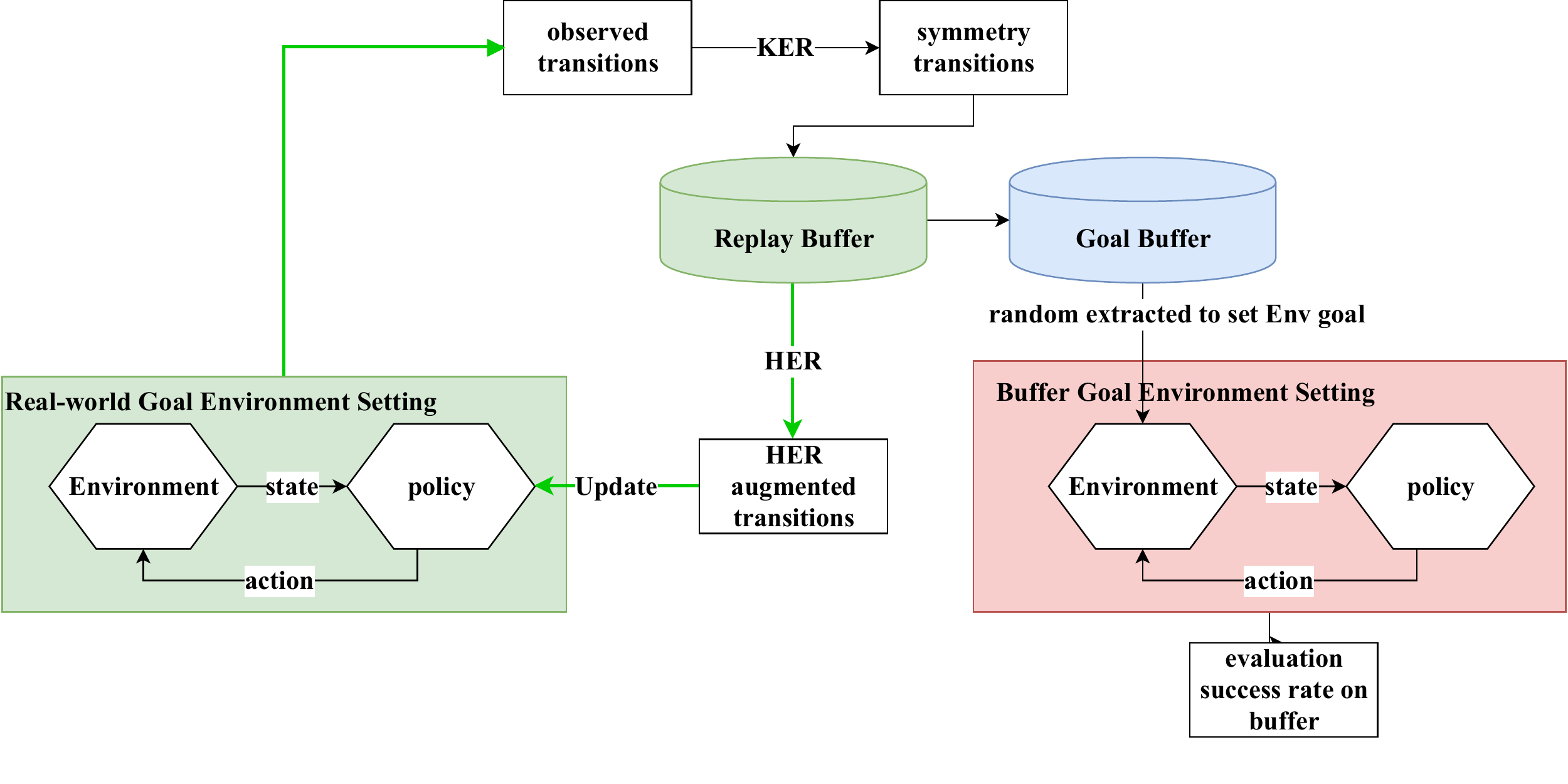}
    \caption{Proposed evaluation framework of the learning performance in terms of buffer data in testing phase}
  \label{fig:diagram_overfit}
\end{figure*}

\begin{figure}
  \centering
        \includegraphics[width=.85\linewidth]{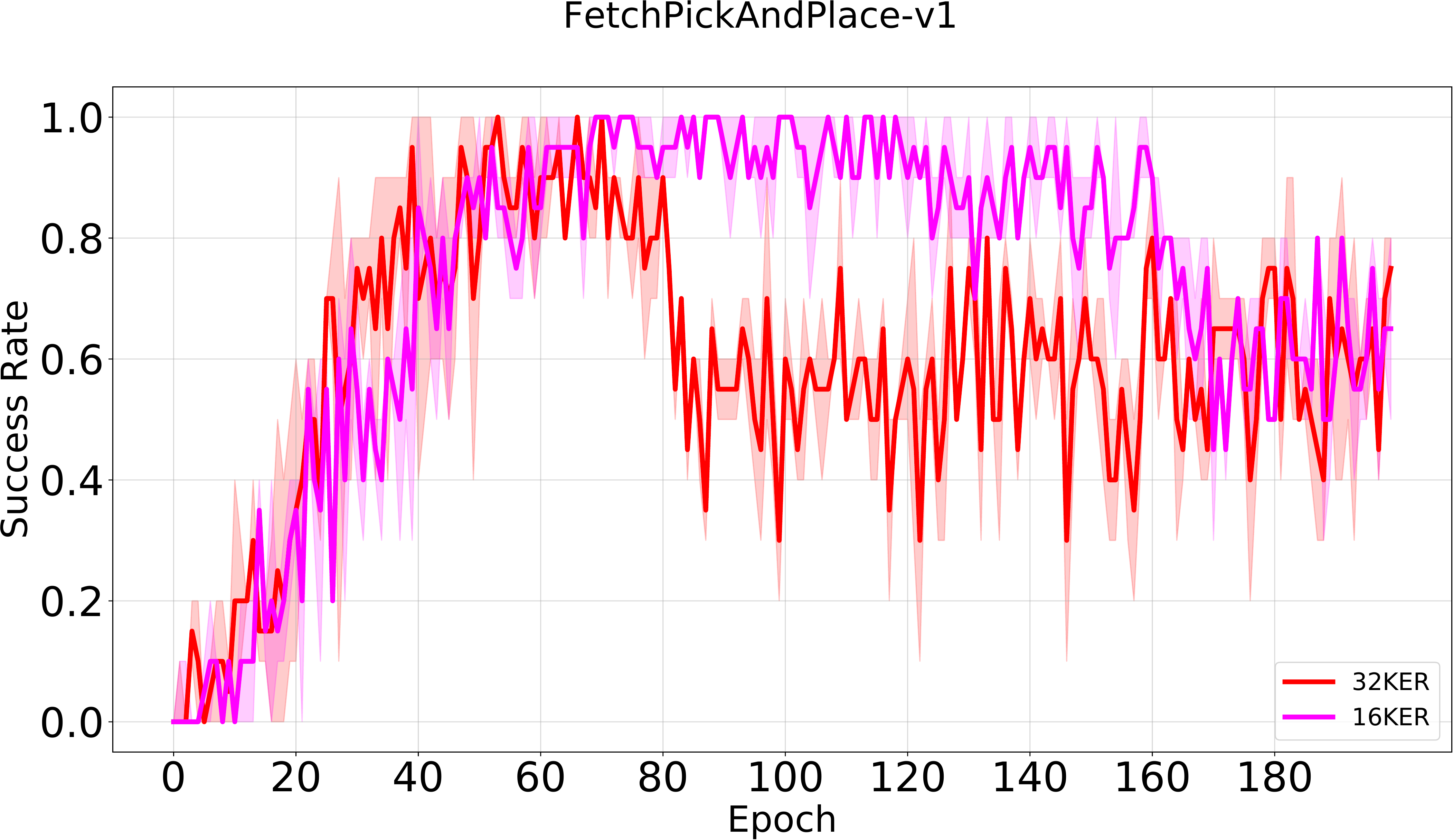}
    \caption{Plots of the learning performance in terms of buffer data in testing phase}
  \label{fig:plot_overfit_pnp}
\end{figure}

\begin{figure}
  \centering
        \includegraphics[width=.85\linewidth]{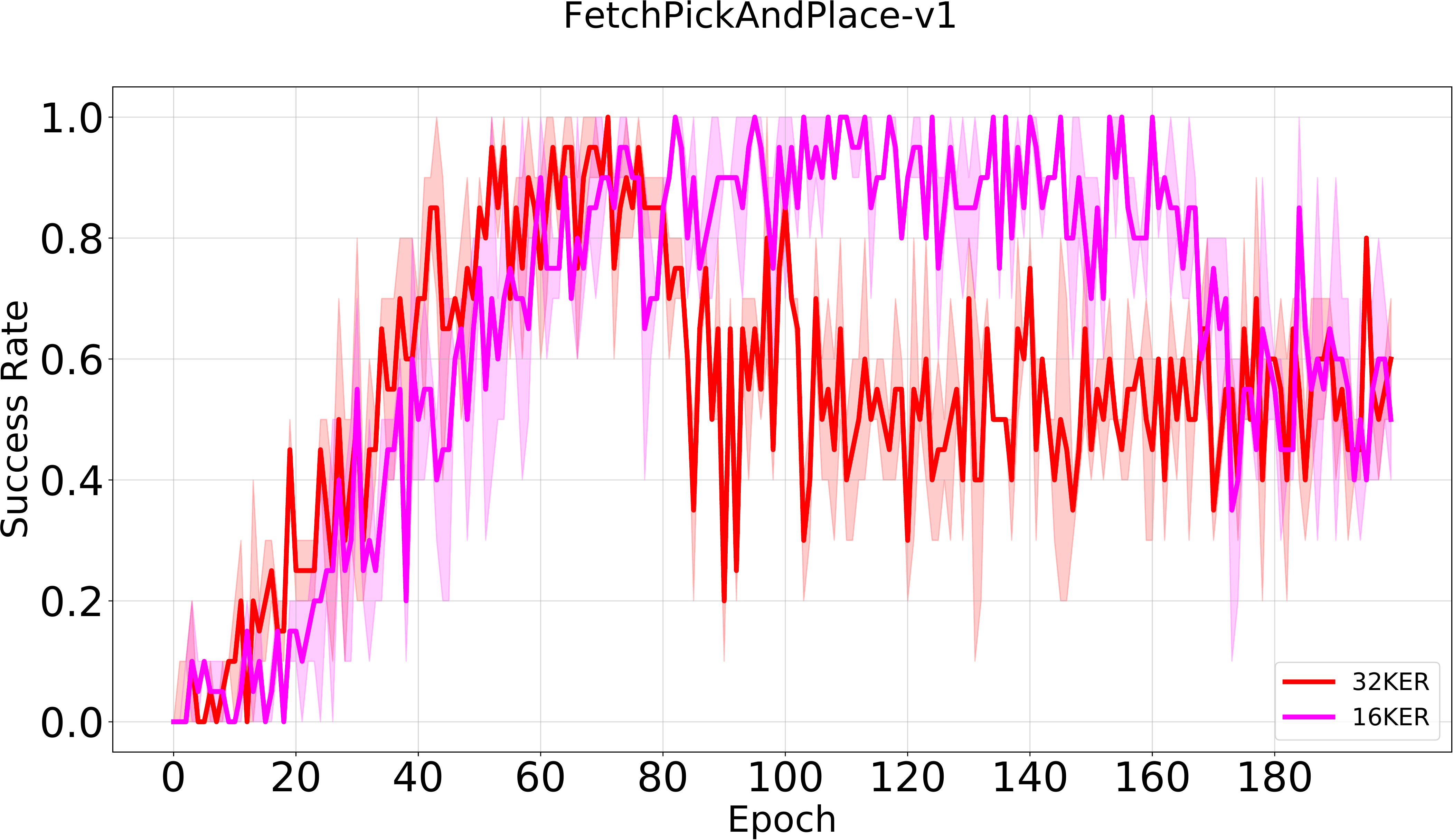}
    \caption{Plots of the learning performance with stopping using HER in 40 epochs}
  \label{fig:terminate_her}
\end{figure}

\item {\bfseries Hypothesis 2: Data Augmentation or DDPG caused the Performance Drop}\\
In those experiments met with performance drop, we used KER, HER and DDPG to train policies. Does any of these techniques cause the drop? To answer this question, first, we tested and observed that what happen when we do not use HER in around 40 epochs (since KER could not learn without her, we used HER until the robot starts to learn), with $n_{ker}=16$ and $n_{ker}=32$ through all the training. The experimental result is shown in Fig. \ref{fig:terminate_her}, which is still met with the drop, proving that it is not HER caused the drop. Then we conducted long-run trains only with HER and DDPG, and the drops are also showed up around 4700 epochs (around 23,500,000 time steps) in Fig. \ref{fig:ddpg_her_long_run} (shadow color may not show in a browser view, please download it and read it offline). These experiments sufficiently proved that the drop is caused by the instability of DDPG, instead of KER nor HER (since learning only with DDPG can never succeed, here we do not include this experiment to help to prove this claim).

\begin{figure}
  \centering
        \includegraphics[width=1.\linewidth]{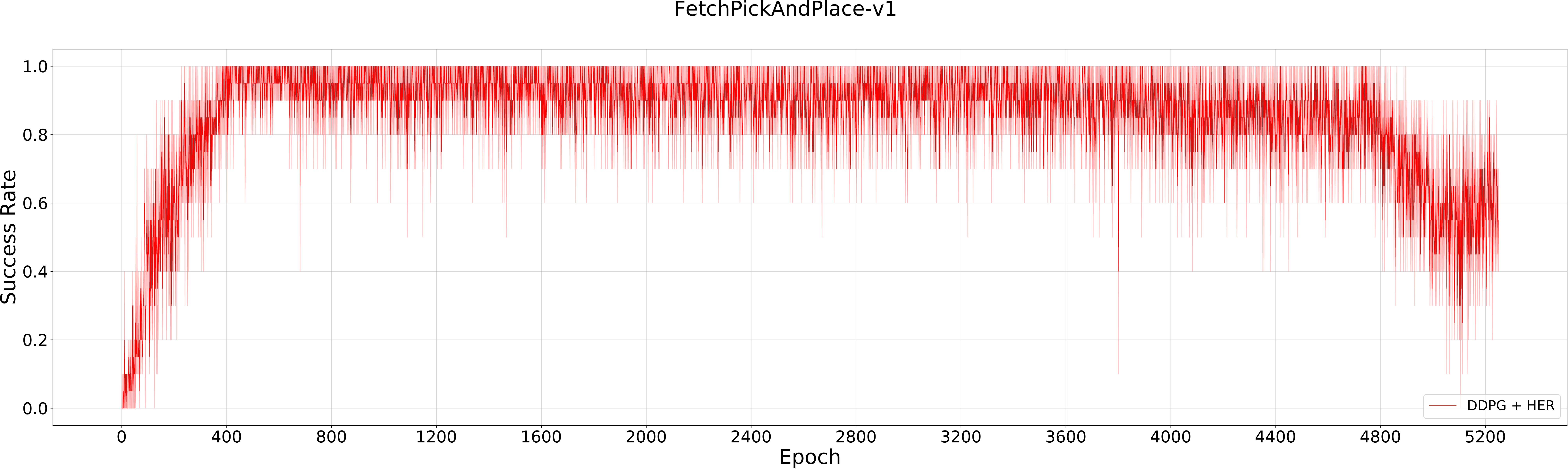}
    \caption{Plots of the learning performance only with HER and DDPG in testing phase}
  \label{fig:ddpg_her_long_run}
\end{figure}
\end{itemize}

\section{Deployment on the Real Robot}\label{appendix:real_robot}

In this section, we show that a policy trained with ITER can be transferred to a real robot and also achieve satisfied performance.

\begin{figure}[h]
  \centering
        \includegraphics[width=.85\linewidth]{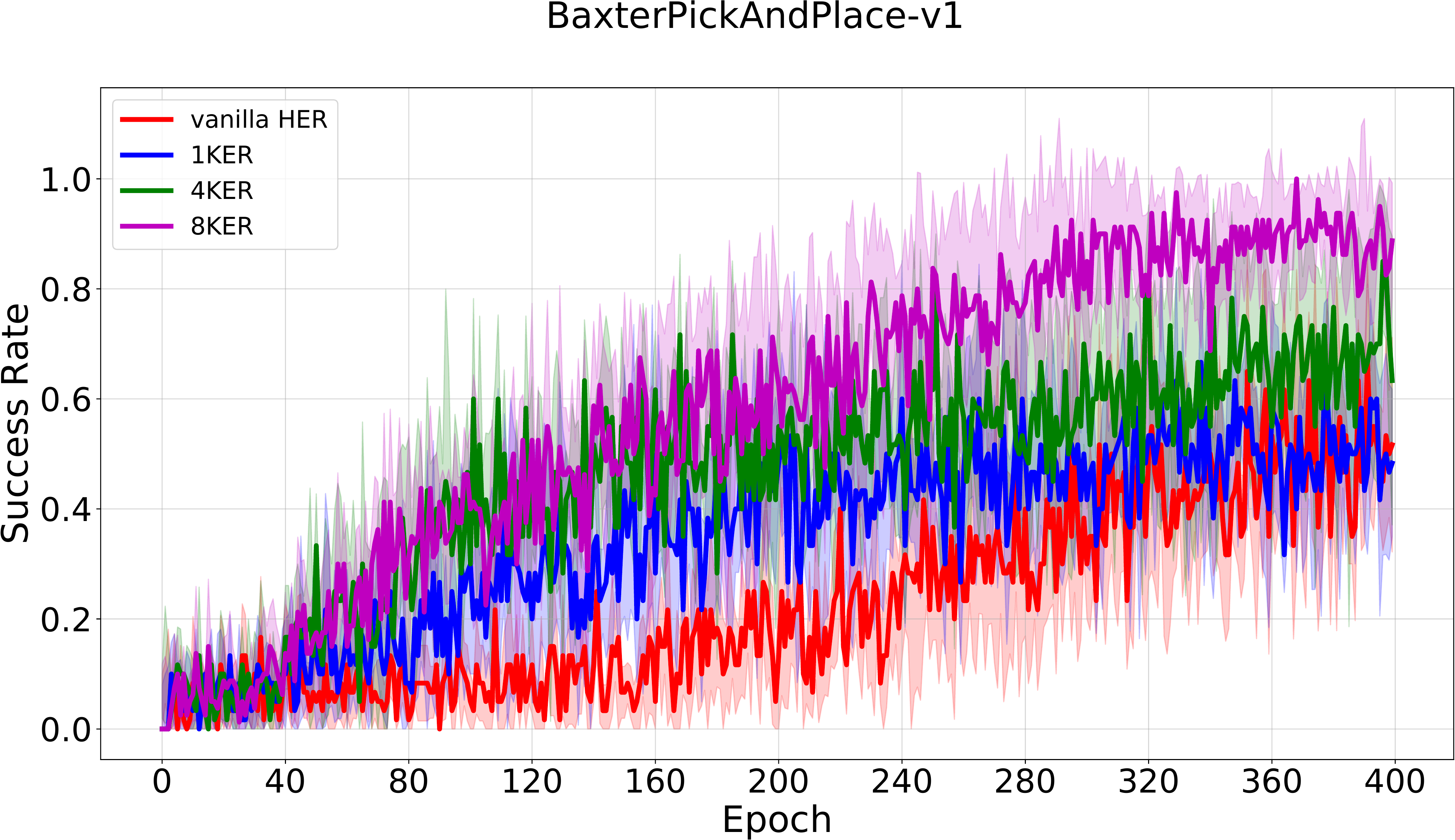}
    \caption{Comparison of different $n_{\text{KER}}$ for KER with a single GER on pick-and-place task with a simulated Baxter}
  \label{fig:baxter_ker}
\end{figure}

\begin{figure}[h]
  \centering
        \includegraphics[width=.85\linewidth]{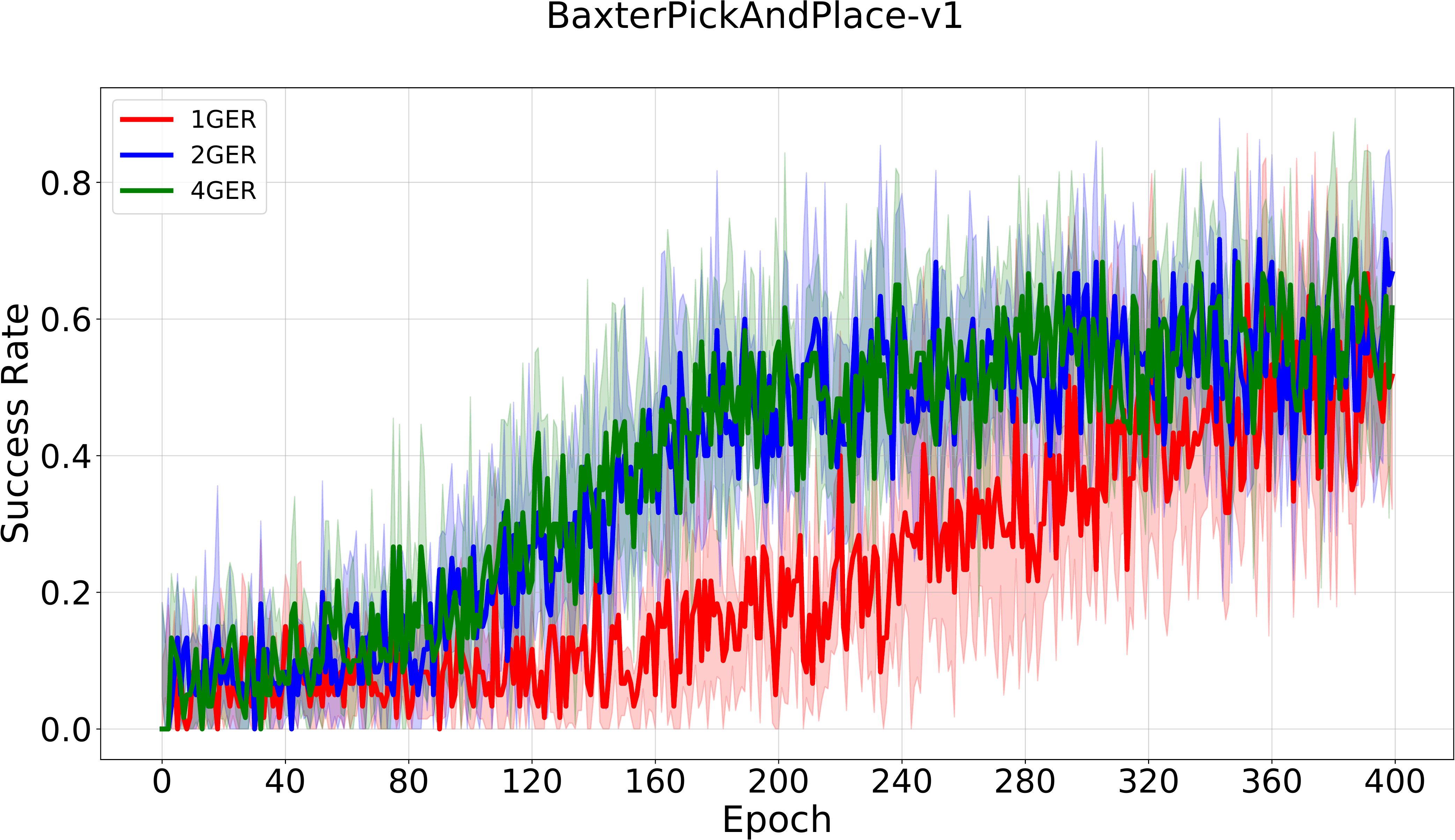}
    \caption{Comparison of different $n_{\text{GER}}$ for GER without KER on pick-and-place task with a simulated Baxter}
  \label{fig:baxter_ger}
\end{figure}

\begin{figure*}[h]
  \centering
        \includegraphics[width=.7\linewidth]{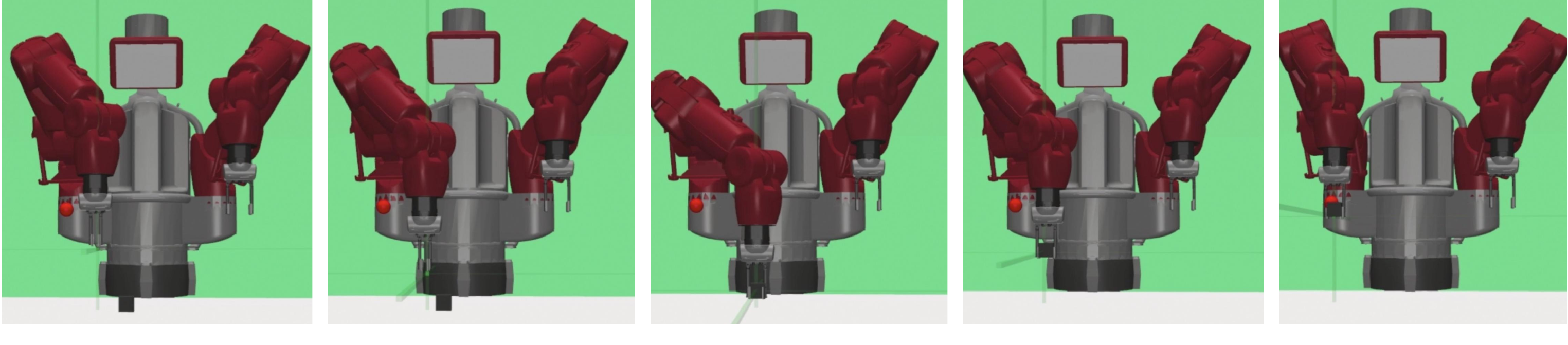}
    \caption{Learned behaviors of simulated Baxter with ITER in pick-and-place task}
  \label{fig:virtual_baxter}
\end{figure*}

To apply a policy to control the real Baxter robot, first we need to train a simulated one in MuJoCo\footnote{Our Baxter simulation code is available at \url{https://github.com/huangjiancong1/gym_baxter}.}. The learning plots of ablative experiments on GER and KER are shown in Figs. \ref{fig:baxter_ker} and \ref{fig:baxter_ger} respectively, and the learned behaviors are shown in Fig. \ref{fig:virtual_baxter}. These experiments show that our method still overwhelms the vanilla HER in the Baxter simulation.

\begin{figure*}[h]
  \centering
        \includegraphics[width=.6\linewidth]{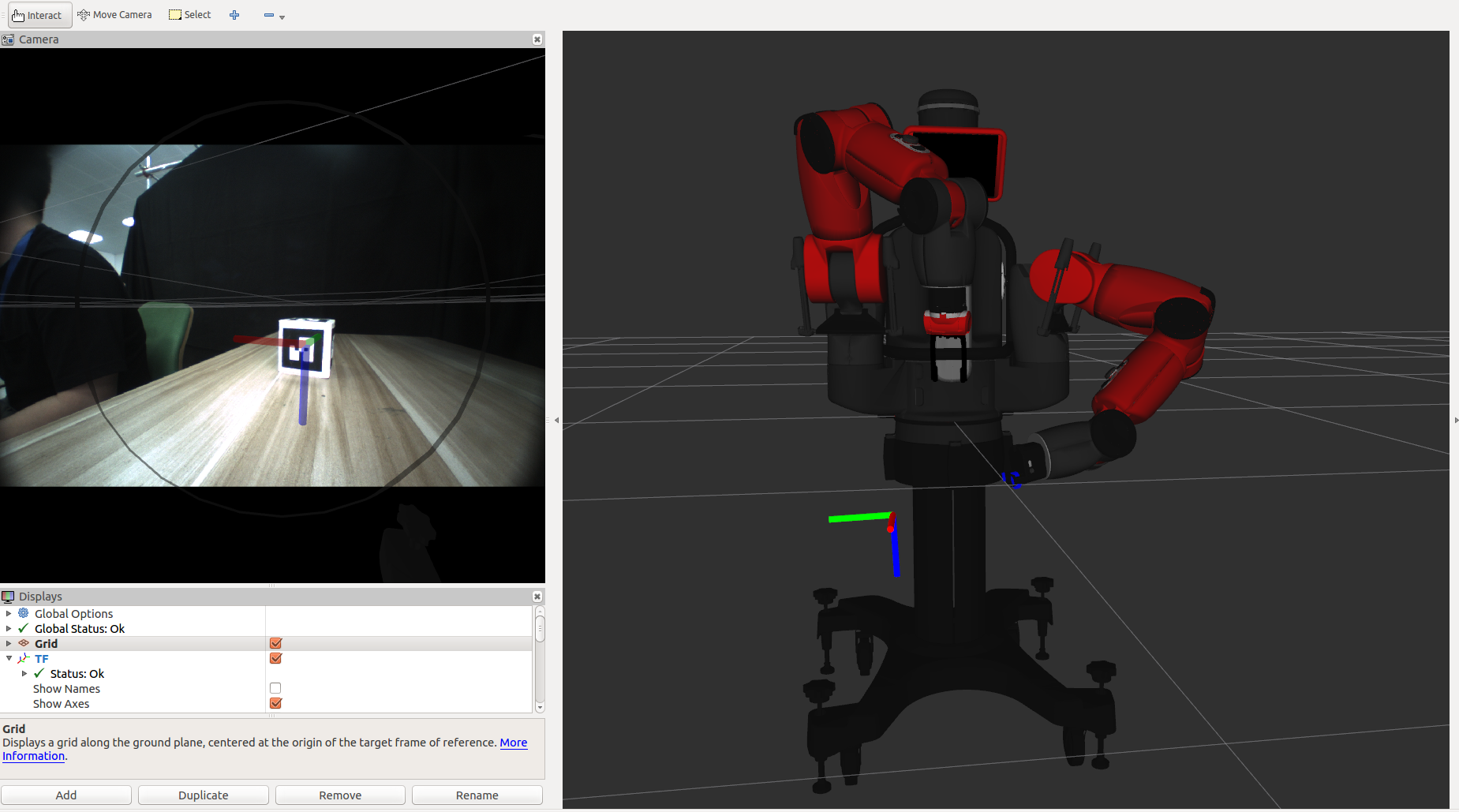}
    \caption{Object pose detection with Alvar markers}
  \label{fig:alvar_RVIZ}
\end{figure*}

\begin{figure*}[t]
\begin{multicols}{2}
\centering
    \includegraphics[width=.5\linewidth]{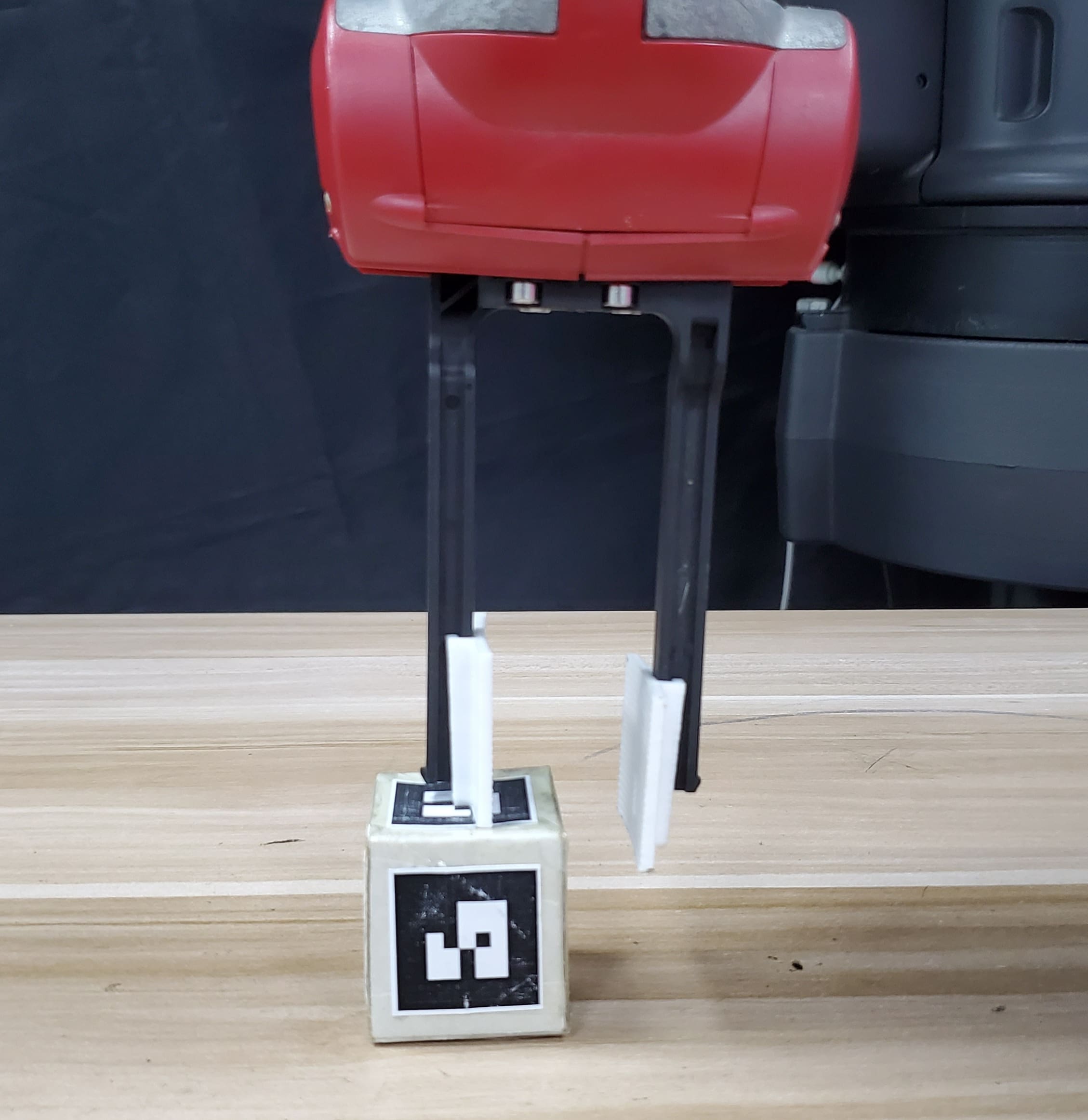}\par 
    \includegraphics[width=.52\linewidth]{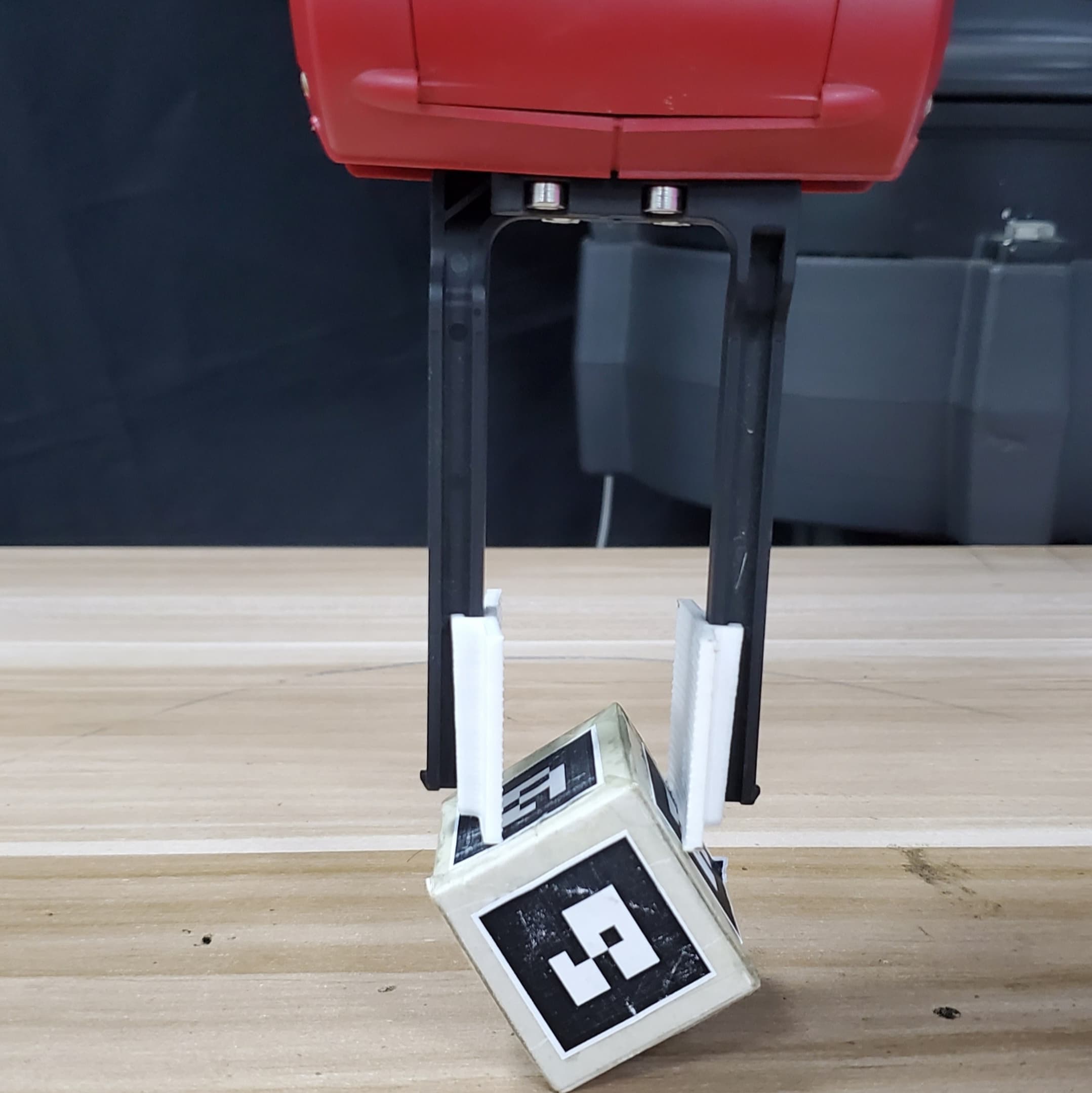}\par 
\end{multicols}
\caption{Failures in the real-world pick-and-place task}
 \label{fig:collision}
\end{figure*}

Then we applied the well-trained policy to the real Baxter as a high-level controller through Robot Operation System (ROS). ROS provides communications between services (\eg the policy) and clients (\eg Baxter). We used Alvar markers\footnote{An open source AR tag tracking library \url{http://wiki.ros.org/ar_track_alvar/}.} to detect the object position (Fig. \ref{fig:alvar_RVIZ}) with a camera equipped in Baxter right arm end effector. To begin an episode, the observed state in real world was sent to the simulator for a virtual environment initialization, and then we keep the policy being used inside. At each timestep, the joints configuration of the simulated Baxter was sent to the real one controlled with position loop (Fig. \ref{fig:exp_real_baxter}). The robot failed in picking up the object and colliding with it (Fig. \ref{fig:collision}) when there is an unexpected shift on the detected pose.  Video is available in \cite{ITER_supplement}.


\section{Experiments on the Single-goal Tasks}\label{appendix:single_goal}
\begin{figure*}[h]
  \centering
    \includegraphics[width=.7\textwidth]{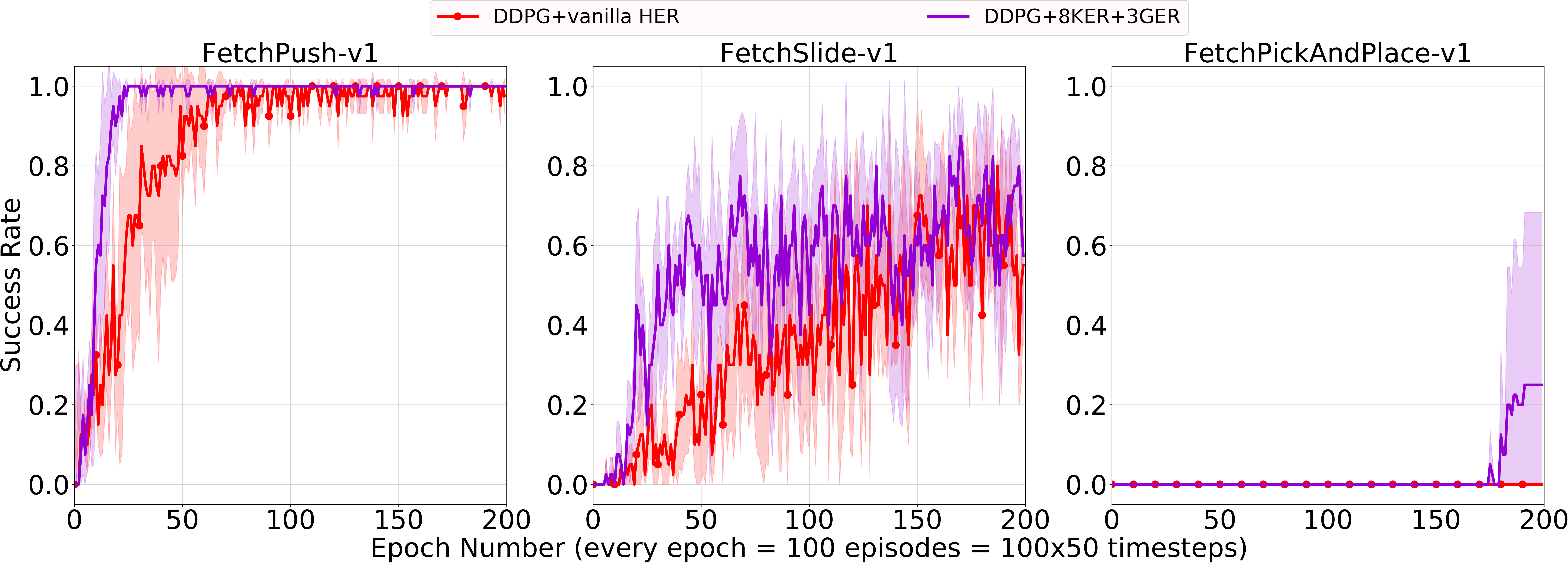}
\caption{Comparison of vanilla HER and ITER with 8 KER symmetries and 4 GER applications on single-goal tasks}
\label{fig:exp_best_singlegoal}
\end{figure*}

For single-goals, our method shows 3 and 4 times speedup over HER in pushing and sliding tasks respectively (Fig. \ref{fig:exp_best_singlegoal}). Surprisingly, our method achieves a starting learning in the pick-and-place task while HER cannot resolve it within 200 epochs (all comparisons are made by measuring the number of epochs to get to convergence of success rates).
The pick-and-place task in the single-goal setting is difficult. Note that in the HER paper, HER does not learn anything without training tricks. Our approach, on the other hand, starts to learn after about 180 epochs.

\end{document}